\documentclass{article}

\usepackage{microtype}
\usepackage{graphicx}
\usepackage{subcaption}
\usepackage{booktabs} 

\usepackage{hyperref}
\usepackage{comment}

\usepackage[accepted]{icml2026}

\usepackage{amsmath}
\usepackage{amssymb}
\usepackage{mathtools}
\usepackage{amsthm}

\usepackage[capitalize,noabbrev]{cleveref}

\theoremstyle{plain}
\newtheorem{theorem}{Theorem}[section]
\newtheorem{proposition}[theorem]{Proposition}

\theoremstyle{definition}
\newtheorem{definition}[theorem]{Definition}

\theoremstyle{remark}

\usepackage[textsize=tiny]{todonotes}

\usepackage{soul}

\usepackage{pifont}
\usepackage{makecell}
\usepackage{multirow}
\newcommand{\cmark}{\ding{51}}  

\usepackage{wrapfig} 
\usepackage{booktabs}
\usepackage{graphicx}

\usepackage[table]{xcolor}
\usepackage[dvipsnames]{xcolor}
\definecolor{green(pigment)}{rgb}{0.0, 0.65, 0.31}
\definecolor{brandeisblue}{rgb}{0.0, 0.44, 1.0}
\definecolor{deepmagenta}{rgb}{0.8, 0.0, 0.8}
\definecolor{orange(colorwheel)}{rgb}{1.0, 0.5, 0.0}

\newcommand{\mosei}[2]{\cellcolor{OrangeRed!#1} #2}
\newcommand{\mosi}[2]{\cellcolor{brandeisblue!#1} #2}
\newcommand{\urfunny}[2]{\cellcolor{orange(colorwheel)!#1} #2}
\newcommand{\mustard}[2]{\cellcolor{green(pigment)!#1} #2}

\usepackage{array}
\newcolumntype{L}[1]{>{\raggedright\let\newline\\\arraybackslash\hspace{0pt}}m{#1}}
\newcolumntype{C}[1]{>{\centering\let\newline\\\arraybackslash\hspace{0pt}}m{#1}}
\newcolumntype{R}[1]{>{\raggedleft\let\newline\\\arraybackslash\hspace{0pt}}m{#1}}

\icmltitlerunning{Toward Structural Multimodal Representations: Specialization, Selection, and Sparsification via Mixture-of-Experts}

\begin{document}

\twocolumn[
  \icmltitle{Toward Structural Multimodal Representations: Specialization, Selection, and Sparsification via Mixture-of-Experts}




  \begin{icmlauthorlist}
    \icmlauthor{Hahyeon Choi}{yyy}
    \icmlauthor{Nojun Kwak}{yyy}
  \end{icmlauthorlist}

  \icmlaffiliation{yyy}{Seoul National University, Seoul, South Korea}

  \icmlcorrespondingauthor{Hahyeon Choi}{hahyeon.choi@snu.ac.kr}
  \icmlcorrespondingauthor{Nojun Kwak}{nojunk@snu.ac.kr}

  \icmlkeywords{Machine Learning, ICML}

  \vskip 0.3in
]



\printAffiliationsAndNotice{}  

\begin{abstract}
We propose S3 (Specialization, Selection, Sparsification), a framework that rethinks multimodal learning through a structural perspective. Instead of encoding all signals into a fixed embedding, S3 decomposes multimodal inputs into semantic experts and selectively routes them for each task. \textit{Specialization} forms concept-level experts in a shared latent space, \textit{Selection} adapts routing for task-specific needs, and \textit{Sparsification} prunes low-utility paths to yield compact, information-minimal representations. Across four MultiBench benchmarks, S3 improves accuracy and shows a consistent reverse U-shaped sparsity–performance trend, with peak performance at intermediate sparsity. These results suggest that structuring multimodal representations as selectable semantic components provides a practical and principled alternative to contrastive learning or InfoMax-driven approaches.
\end{abstract}
\vspace{-0.7cm}
\section{Introduction}
\label{sec:intro}
\vspace{-0.1cm}

Multimodal data capture the same event through different perceptual channels, providing rich and complementary signals~\cite{schrodi2025two, liao2025multimodal, liang2022mind}. However, the information from each modality is inherently asymmetric, as they differ in resolution, coverage, and noise characteristics. Moreover, the semantic overlap between modalities is context-dependent. Each paired sample, therefore, consists of a unique combination of shared and modality-unique information. While humans naturally integrate such heterogeneous signals~\cite{ERNST2004162, GHAZANFAR2006278}, learning algorithms must explicitly determine what to align, what to retain, and what to discard, making multimodal representation learning (MMRL) fundamentally challenging~\cite{locatello2019challenging, tishby2015deep, yu2026latent}.
\vspace{-0.15cm}

Most existing MMRL approaches fall into two dominant paradigms. Contrastive learning methods align paired modalities by projecting them into a shared representation space~\cite{girdhar2023imagebind, jia2021scaling}. While effective at learning stable shared embeddings, this often comes at the cost of discarding task-relevant but modality-unique cues~\cite{liang2023factorized, wang2025an}. The second approach, known as InfoMax-style~\cite{36} methods, aim to preserve all information~\cite{liu2023focal, pan2021disentangled, 9523016}. However, this often results in redundant representations cluttered with task-irrelevant information~\cite{49568, soattoC16}. 
We argue that these limitations stem not only from suboptimal objectives but also from a lack of structural inductive biases. While semantic information is heterogeneously distributed across modalities, most models collapse it into a single, uniform representation, failing to adaptively capture task-relevant information and discard irrelevant variability.

\vspace{-0.10cm}  

To address this, we propose a structural perspective on MMRL. Rather than refining objectives alone, we explicitly decompose multimodal inputs into interpretable semantic units. This idea draws inspiration from recent reinterpretations of Mixture-of-Experts (MoE)~\cite{park2025monet, yang2025mixture}, which treat experts not merely as tools for parameter scaling, but as semantically specialized components that enhance model interpretability and manipulability. Building on this view, we associate each expert in our model with a distinct concept, dynamically composed based on task demands.
Such structured representation offers fine-grained control over what information to retain or discard. Moreover, by grounding alignment in experts consistently activated by shared concepts across data samples, the model can effectively mitigate cross-modal asymmetry while preserving modality-specific nuances.

\vspace{-0.10cm}
Building on this structural view, we introduce S3 (Specialization, Selection, Sparsification), a framework that restructures MMRL into three stages.
(1) Specialization decomposes multimodal signals into concept-level experts, enabling distributed alignment within a structured semantic space.
This representation mitigates cross-modal asymmetry and provides a semantic basis for task-specific selection. 
(2) Selection subsequently navigates this fixed latent space by adapting the routing mechanism, activating task-relevant experts to construct compact, task-focused representations. (3) Sparsification further refines the model at inference time. By pruning low-contribution routing paths, it yields Information-Minimal yet Task-Sufficient representations without additional training.

\vspace{-0.10cm}
Extensive experiments show that S3 consistently outperforms prior methods and maintains structural consistency across four MultiBench~\cite{liang2021multibench} benchmarks.
We find that its routing and sparsification behaviors follow predictable patterns shaped by semantic granularity. Notably, we observe that the sparsification process reveals a reverse U-shaped trend in performance: accuracy peaks at an optimal sparsity level that suppresses task-irrelevant noise while preserving essential signals. These findings support our central hypothesis: structuring multimodal representations as selectable semantic components enables more controllable and efficient information use. S3 thus offers a practical and theoretically grounded path toward \textit{Task-Sufficient} and \textit{Information-Minimal} MMRL.
\vspace{-0.3cm}
\section{A Fundamental Limitation of Existing Multimodal Representation Learning}
\label{sec:problem_formulation}
\vspace{-0.1cm}
We consider the self-supervised multimodal setting to analyze limitations of prior MMRL methods. For clarity, we focus on two modalities, denoted by the set ${\cal M} = \{1,2\}$, although the formulation and results naturally extend to an arbitrary number of modalities.
Let $X^m \sim p_{X^m}(\cdot)$ denote the input for modality $m \in \mathcal{M}$. Each encoder $f^m$ maps the input to a stochastic latent representation $Z^m = f^m(X^m)$.
Since the downstream task label $Y$ is inaccessible during training, the learning process follows the Markov chain:
\vspace{-0.1cm}
\begin{equation}
    Y \;\to\; (X^1,X^2) \;\to\; (Z^1, Z^2).
    \label{eq:multimodal_markov_chain}
\end{equation}
\vspace{-0.65cm}
\begin{proposition}[Data Processing Inequality]
\label{prop:dpi_sufficiency}
If $(X^1, X^2)$, $(Z^1, Z^2)$, and $Y$ satisfy the Markov chain in \Cref{eq:multimodal_markov_chain}, then the Data Processing Inequality (DPI) implies:
    \begin{equation} 
        I(X^1, X^2; Y) \ge I(Z^1, Z^2; Y). 
    \end{equation}
Equality holds if and only if $(Z^1, Z^2)$ constitutes a sufficient statistic for $Y$~\cite{soattoC16}.
\end{proposition}
\vspace{-0.1cm}
Thus, once any $Y$-relevant information is lost during encoding, it cannot be recovered by downstream predictors, making Bayes-optimal performance unattainable. A detailed proof of Proposition~\ref{prop:dpi_sufficiency} is provided in \cref{app:proof_dpi_sufficiency}.
%
%
\begin{figure}
    \centering
    \includegraphics[width=0.84\linewidth]{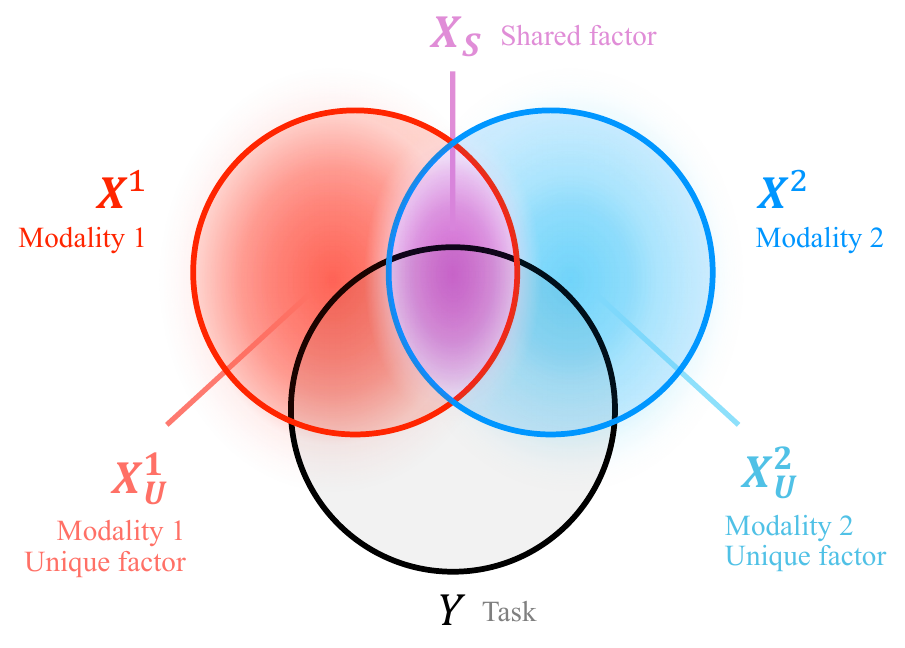}
    \vspace{-0.1cm}
    \caption{Latent factor decomposition of multimodal inputs into shared factor $X_S$ and modality-unique factors $(X^1_U, X^2_U)$. The downstream task $Y$ may depend on some subset of them.
    }
    \vspace{-0.4cm}
    \label{fig:latent_factor_model}
\end{figure}
%
%
\vspace{-0.7cm}
\subsection{Contrastive Learning: Modality-Aligned but Shared-Only Representations}
\label{subsec:limitation_of_multimodal_CL}
\vspace{-0.15cm}
Multimodal contrastive learning promotes cross-modal alignment by maximizing the mutual information between paired modalities~\cite{almudevar2025aligning, arandjelovic2017look}.
Let $f_{\text{CL}}^m$ be the contrastively-learned encoder, producing a stochastic representation $Z_{\text{CL}}^m = f_{\text{CL}}^m(X^m) $, whose objective can be written as:
\vspace{-0.1cm}
\begin{equation}
    \max_{f_{\text{CL}}^1,\, f_{\text{CL}}^2}
    \left[ I(Z_{\text{CL}}^1 ; X^2) + I(X^1 ; Z_{\text{CL}}^2) \right].
    \label{eq:cl_objective}
\end{equation}
While this objective encourages the encoders to capture cross-modal dependencies, it systematically suppresses task-relevant signals when the downstream task depends on modality-unique cues.
To formalize this limitation, we model the latent structure of multimodal inputs using the following factorization.
\begin{definition}[Multimodal Latent Factor Model]
\label{def:latent_factor_model}
    Each modality input $X^m$ consists of a shared factor $X_S$ and a modality-unique factor $X_U^m$:
    \vspace{-0.1cm}
    \begin{equation}
        X^1 = (X_S, X_U^1),
        \qquad
        X^2 = (X_S, X_U^2),
    \end{equation}
    and these factors satisfy the independence conditions:
    \vspace{-0.1cm}
    \begin{equation}
        X_S \perp X_U^1,
        \quad
        X_S \perp X_U^2,
        \quad
        X_U^1 \perp X_U^2.
    \label{eq:independence}
    \end{equation}
\end{definition}
\vspace{-0.2cm}
This structure is illustrated in \cref{fig:latent_factor_model}, which depicts the decomposition into shared and modality-unique factors.
\begin{proposition}[Mutual Information Decomposition]
\label{prop:mi_decomposition}
    Under Definition~\ref{def:latent_factor_model}, the mutual information between modalities satisfies
    \vspace{-0.1cm}
    \begin{equation}
    I(X^1 ; X^2) = H(X_S),
    \end{equation}
    where $H(\cdot)$ denotes the entropy. In other words, the modality-unique factors $X_U^1$ and $X_U^2$ make no contribution to the mutual information between the modalities.
\end{proposition}
\vspace{-0.2cm}
By Proposition~\ref{prop:mi_decomposition}, the theoretical optimum of the contrastive objective depends only on the shared factor $X_S$. This optimal structure offers no theoretical incentive to preserve modality-unique information $X_U^m$. Consequently, the contrastive representations $Z_{\text{CL}}^m$ are effectively characterized by the following Markov chain:
\vspace{-0.1cm}
\begin{equation}
    Y 
    \; \to \; (X_S, X_U^1, X_U^2)
    \; \to \; X_S
    \; \to \; (Z_{\text{CL}}^1, Z_{\text{CL}}^2).
    \label{eq:cl_markov}
\end{equation}
We now consider the downstream setting in which the task depends on modality-unique factors.
\begin{definition}[Task Relevance of Unique Factors]
    \label{def:task_unique}
    We say that the downstream task $Y$ depends on modality-unique factors when the following condition holds:
    \vspace{-0.1cm}
    \begin{equation}
        I(X_U^1, X_U^2 ; Y) > 0.
    \end{equation}
\end{definition}
\vspace{-0.3cm}
\begin{proposition}[Fundamental Limitation of Contrastive Representations]
\label{prop:cl_limitation}
If the condition in Definition~\ref{def:task_unique} holds, then the contrastively-learned representations satisfy:
    \vspace{-0.1cm}
    \begin{equation}
    I(X^1, X^2; Y)
    >
    I(Z_\text{{CL}}^1, Z_{\text{CL}}^2; Y).
    \end{equation}
\end{proposition}
\vspace{-0.3cm}
This strict inequality follows directly from Proposition~\ref{prop:dpi_sufficiency}. It implies that $Z_{\text{CL}}^m$ cannot serve as a sufficient statistic when the task depends on modality-unique features.
Despite strong cross-modal alignment, the contrastive objective lacks an explicit mechanism to retain modality-unique features, resulting in an inevitable loss of task-relevant information.
Proofs are provided in \Cref{app:proof_mi_decomposition,app:proof_cl_limitation}.
\vspace{-0.6cm}
\subsection{InfoMax: Preserving Everything, Losing Focus}
\label{subsec:limitations_of_infomax}
\vspace{-0.1cm}
As discussed in \Cref{subsec:limitation_of_multimodal_CL}, contrastive learning aligns shared information but overlooks modality-unique factors critical for certain tasks.
To address this limitation, recent works~\cite{wang2025an, liu2023focal, pan2021disentangled} adopt InfoMax-style approaches~\cite{36, hjelm2018learning} that aim to preserve both shared and modality-unique information.
In this setting, each input $X^m$ is encoded into a shared representation $Z_S$ and a modality-unique representation $Z_U^m$ via encoders $f_{\text{max}}^m$, modeled as $(Z_S, Z_U^m) = f^m_{\text{max}}(X^m)$.
These methods typically optimize
\vspace{-0.1cm}
\begin{equation}
    \max_{f_\text{max}^1, f_\text{max}^2} \sum_{m} \left[ I(Z_S;X^m) + I(Z_U^m;X^m|Z_S) \right].
    \label{eq:infomax_objective}
\end{equation}
Defining $Z_\text{max}^m = (Z_S, Z_U^m)$, the objective becomes, by the chain rule, equivalent to maximizing $\sum_m I(Z_\text{max}^m; X^m)$, which encourages the model to retain all information from the input. As a result, InfoMax is possible to satisfy the equality condition in Proposition~\ref{prop:dpi_sufficiency} and aims to produce a $Y$-sufficient representation. However, it also inherently retains substantial task-irrelevant information.
\begin{definition}[Task-Irrelevant Information]   
        The task-irrelevant information captured by representation $Z^m$ from $X^m$, conditioned on $Y$, is defined as:     
    \vspace{-0.1cm}
    \begin{equation} 
        I(Z^m ; X^m | Y). 
    \end{equation} 
    \label{def:irrelevant_information}
\end{definition}
\vspace{-0.9cm}
Under the Markov chain in \cref{eq:multimodal_markov_chain}, the InfoMax objective decomposes as:
\vspace{-0.1cm}
\begin{equation}
    \max_{f^1_\text{max}, f^2_\text{max}} \sum_m \left[ I(Z^m_\text{max}; Y) + I(Z^m_\text{max}; X^m | Y) \right]. 
    \label{eq:infomax_decomposition}
\end{equation}
This decomposition reveals that InfoMax simultaneously maximizes both task-relevant ($I(Z^m_\text{max}; Y)$) and task-irrelevant ($I(Z^m_\text{max}; X^m | Y)$) information. Retaining such irrelevant information has been widely reported as detrimental to downstream performance~\cite{liang2023factorized, soattoC16}.
In particular, \citet{49568} introduced the InfoMin principle in the unimodal setting, emphasizing that representations should be sufficient for the task but minimal with respect to the input.
In summary, while InfoMax-based approaches alleviate the shared-only bias of contrastive learning, they remain suboptimal due to their inability to suppress task-irrelevant information.
\vspace{-0.3cm}
\section{Towards Structurally Optimal Multimodal Representations}
\label{sec:optimal_representation}
\vspace{-0.1cm}
\subsection{Task-Sufficiency and Information-Minimality}
\label{subsec:optimal_representation}
\vspace{-0.1cm}
As discussed in \cref{sec:problem_formulation}, existing approaches fail in complementary ways. Contrastive learning discards modality-unique signals essential for certain tasks, whereas InfoMax-style methods preserve excessive task-irrelevant information.
This motivates the need to precisely characterize the balance between information retention and suppression in multimodal settings.
To this end, we extend the InfoMin principle~\cite{49568} to multimodal settings and formalize the conditions that an optimal task-specific representation $(Z_Y^{1*}, Z_Y^{2*})$ must satisfy. 
Under the Markov chain in \cref{eq:multimodal_markov_chain}, such representations must preserve all information relevant to the target $Y$.
This corresponds to the equality condition in Proposition~\ref{prop:dpi_sufficiency}.
\begin{definition}[Task-Sufficiency]
    $(Z_Y^{1*}, Z_Y^{2*})$ satisfies Task-Sufficiency if~and~only~if it is a sufficient statistic~for~$Y$:
    \vspace{-0.1cm}
    \begin{equation}
        I(Z_Y^{1*}, Z_Y^{2*} ; Y) = I(X^1, X^2;Y).
    \end{equation}
\label{def:task-sufficiency}
\end{definition}
\vspace{-0.7cm}
However, Task-Sufficiency alone does not ensure the exclusion of irrelevant factors.
To be optimal, representations must exclude all information that is independent of $Y$.
\begin{definition}[Information-Minimality]
    $(Z_Y^{1*}, Z_Y^{2*})$ satisfies Information-Minimality if it is conditionally independent of the inputs $(X^1, X^2)$ given $Y$:
    \vspace{-0.1cm}
    \begin{equation}
        I( Z_Y^{1*}, Z_Y^{2*}; X^1, X^2 | Y) = 0.
    \end{equation}
\label{def:information-minimality}
\end{definition}
\vspace{-0.8cm}
Together, these two conditions imply that $(Z_Y^{1*}, Z_Y^{2*})$ forms a minimal sufficient statistic for $Y$~\cite{Soatto2014VisualSR}, which we treat as our theoretical optimum.
\vspace{-0.3cm}
\subsection{From Monolithic to Structured Representations}
\vspace{-0.1cm}
While these formulations define an ideal target, achieving such optimality in practice is challenging. Multimodal inputs often consist of complex mixtures of shared and modality-unique factors, whose task-relevance varies across samples. Consequently, a single monolithic embedding cannot flexibly adapt to such variability, limiting the generalization and expressivity of learned representations.

\vspace{-0.1cm}
To overcome this limitation, we shift towards a structured representation that organizes information into semantically meaningful components.
We structure representations based on latent semantic concepts that may appear in one or both modalities.
Instead of aligning raw inputs directly, we aim to align distributions over concept-specific subspaces, which allows for the flexible reuse of shared semantics while accommodating modality-specific nuances.
This perspective offers potential robustness to cross-modal asymmetry and supports more principled generalization.
We formalize this intuition through the notion of Distributional Semantic Coherence (DSC), which forms the foundation of our proposed S3 framework.
\begin{definition}[Latent Semantic Concept Space]
\label{def:latent_semantic_concept}
Let $\mathcal{Z}$ denote the joint latent representation space.
We assume a set of latent semantic concepts $\mathcal C$ and model 
$\mathcal{Z}$ as a structured space composed of concept-wise subspaces $\{\mathcal{Z}_c\}_{c \in \mathcal{C}}$ such that $\mathcal{Z}=\bigoplus_{c \in \mathcal{C}} \mathcal{Z}_c$, where each $c \in \mathcal{C}$ corresponds to a distinct latent semantic concept.
For an input sample $x^m$, a realization of $X^m$, the set of active concepts is defined as
\vspace{-0.1cm}
\begin{equation}
    C^m(x^m)=\{ c \in \mathcal{C} | \, \|\pi_c(f^m(x^m))\| > \epsilon\},
\end{equation}
where $\pi_c : \mathcal{Z} \to \mathcal{Z}_c$
is the concept-wise projection operator
and $\epsilon \ge 0$ is a sparsity threshold.
\end{definition}
\vspace{-0.1cm}
\begin{definition}[Distributional Semantic Coherence]
\label{def:distributional_semantic_coherence_formal}
A concept $c \in \mathcal{C}$ is shareable if
$\Pr(c \in C^1(X^1) \cap C^2(X^2)) > 0$\footnote{Consider ({image}, {audio}) data. If the dataset contains a (\textsc{dog} image, \textsc{dog} sound) pair, the concept of \textsc{dog} is shareable. However, the concept of a color (e.g., \textsc{black}) cannot be shared because audio data cannot contain the color concept.}.  A multimodal representation satisfies DSC if, for every shareable concept $c$, the distributions of its concept-specific latent variables
are identical across modalities:
\vspace{-0.1cm}
\begin{equation}
    p(\pi_c(Z^1) | c \in C^1(X^1))
    =
    p(\pi_c(Z^2) | c \in C^2(X^2)).
    \label{eq:dsc}
\end{equation}
\end{definition}
\vspace{-0.5cm}
\section{Preliminaries}
\label{sec:preliminaries}
\begin{figure}
    \centering
    \includegraphics[width=0.9\linewidth]{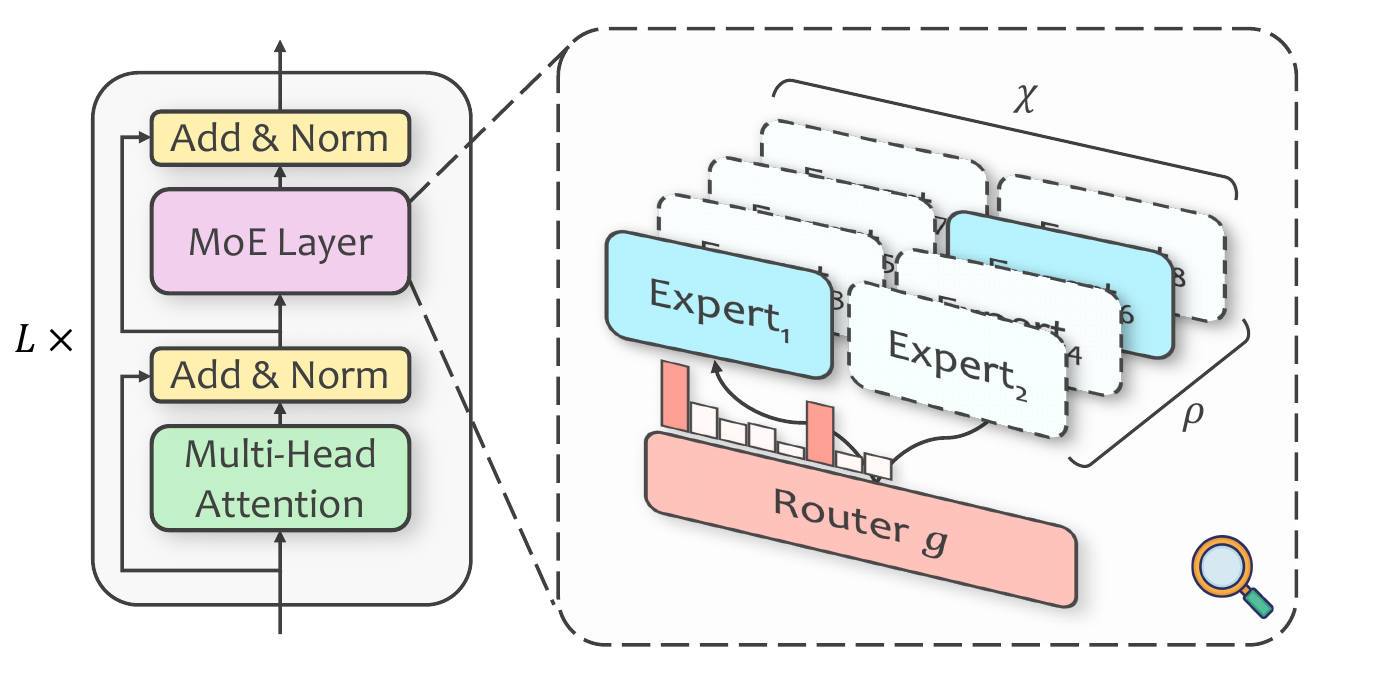}
    \vspace{-0.15cm}
    \caption{Overview of the MoE layer, where a router selects a sparse subset of experts for each input token, and the granularity $\chi$ and expansion ratio $\rho$ determine the expert configuration.
    }
    \vspace{-0.5cm}
    \label{fig:moe}
\end{figure}

\vspace{-0.15cm}
In this section, we summarize the operating principles of the Mixture-of-Experts (MoE) architecture (\cref{fig:moe}) that underpins the proposed S3 framework. For notational simplicity, we omit the modality index $m$ throughout \cref{sec:preliminaries}.
\vspace{-0.65cm}
\subsection{Transformer}
\label{subsec:transformer}
\vspace{-0.1cm}
The Transformer~\cite{vaswani2017attention} consists of $L$ stacked layers, each composed of a Multi-Head Attention (MHA) module followed by a Feed-Forward Network (FFN). We focus our description on the FFN computation, as it serves as the baseline component replaced by the MoE layer.
Given an intermediate token representation ${\mathbf x} \in {\mathbb R}^{D_\text{model}}$ at a particular encoder layer, the FFN applies two linear transformations with a nonlinear activation $\phi$ (e.g., GeLU or ReLU):
\vspace{-0.1cm}
\begin{equation}
    {\rm FFN}({\mathbf x}) = {\mathbf W}_2 \, \phi({\mathbf W}_1 {\mathbf x} + {\mathbf b}_1) + {\mathbf b}_2,
    \label{eq:ffn}
\end{equation}
where $\mathbf{W}_1 \in \mathbb{R}^{D_\text{ffn} \times D_\text{model}}$,
$\mathbf{b}_1 \in \mathbb{R}^{D_\text{ffn}}$, 
$\mathbf{W}_2 \in \mathbb{R}^{D_\text{model} \times D_\text{ffn}}$,
$\mathbf{b}_2 \in \mathbb{R}^{D_\text{model}}$ are the learnable parameters of the FFN. The hidden dimension typically follows $D_\text{ffn} = 4 \cdot D_\text{model}$~\cite{radford2018improving, Rae2021ScalingLM, journals/corr/abs-2302-13971}.
\vspace{-0.4cm}
\subsection{Mixture-of-Experts (MoE)}
\label{subsec:mixture_of_experts}
\vspace{-0.1cm}
The MoE architecture~\cite{shazeer2017} replaces the dense FFN in the Transformer with multiple experts, activating only a small subset per input to enable conditional computation. An MoE layer consists of $N_{\text{expert}}$ independent experts $\{e_i\}_{i=1}^{N_{\text{expert}}}$ and a router $g: \mathbb{R}^{D_{\text{model}}} \rightarrow \mathbb{R}^{N_{\text{expert}}}$. 
Given a token representation $\mathbf x$, the MoE output is defined as:
\vspace{-0.15cm}
\begin{equation}
    {\rm MoE}({\mathbf x}) = \sum_{i=1}^{N_\text{expert}}g({\mathbf x})_i \, e_i({\mathbf x}).
    \label{eq:moe_output}
\end{equation}
Each expert $e_i$ is a two-layer MLP identical in form to the standard FFN:
\vspace{-0.15cm}
\begin{equation}
    e_i({\mathbf x}) = {\mathbf W}_2^i \, \phi({\mathbf W}_1^i {\mathbf x} + {\mathbf b}_1^i) + {\mathbf b}_2^i,
\end{equation}
where the expert parameters are 
$\mathbf{W}_1^{i} \in \mathbb{R}^{D_{\text{expert}} \times D_{\text{model}}}$,
$\mathbf{b}_1^{i} \in \mathbb{R}^{D_{\text{expert}}}$,
$\mathbf{W}_2^{i} \in \mathbb{R}^{D_{\text{model}} \times D_{\text{expert}}}$,
$\mathbf{b}_2^{i} \in \mathbb{R}^{D_{\text{model}}}$. The choices of $D_{\text{expert}}$ and $N_{\text{expert}}$ are discussed in \Cref{subsec:granularity}.
%
%
\vspace{-0.25cm}
\subsection{Router}
\vspace{-0.1cm}
The router $g$ determines which experts are activated by assigning a routing weight $g({\mathbf x})_i$ to each expert $e_i$, conditioned on the input $\mathbf x$. It is typically implemented as a linear projection ${\mathbf W}_g \in {\mathbb R}^{N_\text{expert} \times D_\text{model}}$, and its basic operation is 
\vspace{-0.1cm}
\begin{equation}
    g({\mathbf x}) = {\rm TOP}_k({\rm softmax}({\mathbf W}_g {\mathbf x})).
    \label{eq:router}
\end{equation}
The softmax produces routing scores over all experts, and $\rm{TOP}_k(\cdot)$ selects the top-$k$ experts to enable sparse routing. Thus, the input $\mathbf x$ is forwarded only to these selected experts. The router does not transform the input feature, instead, it outputs a control signal that governs expert selection.
To prevent excessive load concentration on a small subset of experts~\cite{zhou2022mixture,lewis2021base}, auxiliary routing losses are commonly used during training~\cite{lepikhin2021gshard, fedus2022switch}. The specific auxiliary losses adopted in our framework are detailed in \Cref{app:auxiliary_losses}.
\vspace{-0.25cm}
\subsection{Granularity}
\label{subsec:granularity}
\vspace{-0.1cm}
Granularity ($\chi$) and expansion ratio ($\rho$) are hyperparameters that control the structural decomposition of an MoE layer and its overall parameter capacity~\cite{pmlr-v235-ludziejewski24a}. The granularity $\chi$ specifies how the hidden dimension of the FFN, $D_\text{ffn}$, is partitioned across experts:
\vspace{-0.15cm}
\begin{equation}
    \chi = \frac{D_\text{ffn}}{D_\text{expert}}.
\end{equation}
The expansion ratio $\rho$ denotes the parameter increase of the MoE layer relative to a single FFN:
\vspace{-0.15cm}
\begin{equation}
    \rho = \frac{P_\text{moe}}{P_\text{ffn}},
\end{equation}
where $P_\text{moe}$ counts the parameter of all experts (excluding the router), and $P_\text{ffn}$  is that of an FFN with the same architecture. The total number of experts is $N_\text{expert}= \chi \cdot \rho$, determined by granularity and expansion ratio.
\vspace{-0.3cm}
\section{S3: A Modular and Structured Approach to Multimodal Representation Learning}
\label{sec:methods}
\vspace{-0.1cm}
To realize the structured representation principles outlined in \Cref{sec:optimal_representation}, we introduce S3, a three-stage framework for MMRL.
Our goal is to construct representations that are \textit{Task-Sufficient} and \textit{Information-Minimal}, while also satisfying the structural constraints of \textit{latent concept decomposition} and \textit{Distributional Semantic Coherence}.
S3 builds on modality-specific MoE encoders that decompose inputs into expert-organized semantic representations, where each subspace corresponds to a distinct latent concept.
Each expert specializes in a distinct semantic factor, enabling structural disentanglement.
A lightweight router then selectively activates a subset of experts per input by identifying task-relevant semantic components.
The following subsections detail each stage of the S3 framework: Specialization, Selection, and Sparsification.

\vspace{-0.25cm}
\subsection{Expert Pretraining for Specialization}
\label{subsec:pretraining}
\vspace{-0.1cm}
The goal of the Specialization stage is to construct a latent space that serves as the semantic basis for task-adaptive routing. This space should be semantically expressive and structurally coherent, organizing recurring semantic factors across modalities while preserving modality-specific details. 
To this end, we pretrain modality-specific MoE encoders $(f^1, f^2)$ to decompose inputs into concept-level representations distributed across expert-specific subspaces $\{\mathcal{Z}_c\}_{c \in \mathcal{C}}$.
This structure enables the model to disentangle multimodal signals into modular semantic units, forming a compositional basis from which task-relevant information can later be selectively extracted.
\vspace{-0.4cm}
\paragraph{Optimization Objective.}
To realize this structured latent space, we formulate the pretraining objective to maximize semantic coverage within each modality while ensuring Distributional Semantic Coherence across modalities:
\vspace{-0.1cm}
\begin{equation}
    \underset{f^1, f^2}{\max} \boldsymbol{[} \,
        I(Z^1;X^1) + I(Z^2;X^2)
    \, \boldsymbol{]}, 
    \quad \text{s.t. DSC}.
    \label{eq:specialization_objective}
\end{equation}
Here, maximizing mutual information encourages each encoder to preserve rich semantics for each modality, while the DSC constraint in \cref{eq:dsc} enforces concept-wise cross-modal alignment.  
\vspace{-0.4cm}
\paragraph{Training Loss Formulation.} 
Directly computing the mutual information in \cref{eq:specialization_objective} is intractable, so we approximate its variational lower bound using InfoNCE~\cite{journals/corr/abs-1807-03748}. For a sample $x_i^m$, the encoder output is $z_i^m = f^m(x_i^m)$, and for any modality pair $(m, \bar{m}) \in \mathcal{M} \times \mathcal{M}$, the InfoNCE objective is:
\vspace{-0.15cm}
\begin{equation}
    \mathcal{L}_{\text{InfoNCE}}^{[m \rightarrow \bar{m}]} =
    - \mathbb{E}_{i} \left[
        \log \frac{
            \exp\left( \langle z_i^m, z_i^{\bar{m}} \rangle / \tau \right)
        }{
            \sum_{j} \exp\left( \langle z_i^m, z_j^{\bar{m}} \rangle / \tau \right)
        }
    \right],
\end{equation}
where $\langle \cdot, \cdot \rangle$ denotes $\ell_2$-normalized cosine similarity and $\tau$ is the temperature. To satisfy the objective \Cref{eq:specialization_objective}, we employ three loss components.
$\mathcal{L}_\text{rep}$ maximizes $I(X^m;Z^m)$ by encouraging each encoder to preserve rich and diverse semantic factors within its modality:
\vspace{-0.15cm}
\begin{equation}
    {\cal L}_\text{rep} = \frac{1}{2}\left({\cal L}_\text{InfoNCE}^{[1 \rightarrow 1]} + {\cal L}_\text{InfoNCE}^{[2 \rightarrow 2]}\right).
    \label{eq:L_rep}
\end{equation}
The hard DSC constraint is accounted for by $\mathcal{L}_\text{dsc}$, which encourages samples that share the same latent concept to occupy nearby regions across modalities:
\vspace{-0.15cm}
\begin{equation}
    {\cal L}_\text{dsc} = \frac{1}{2} \left( {\cal L}_\text{InfoNCE}^{[1 \rightarrow 2]} + {\cal L}_\text{InfoNCE}^{[2 \rightarrow 1]} \right).
\end{equation}
Although InfoNCE is computed at the instance level, its contrastive signal implicitly shapes the expert-wise activation patterns, promoting distributional alignment of semantically shared concepts across modalities.

\vspace{-0.2cm}
$\mathcal{L}_\text{aux}$ regularizes the router $g$ by encouraging balanced expert utilization across the data distribution and promoting confident, input-conditioned expert activation. This prevents expert collapse and facilitates semantic specialization by distributing distinct features across experts.
A detailed formulation is provided in \Cref{app:auxiliary_losses}. The final training objective for Specialization is the weighted sum:
\vspace{-0.15cm}
\begin{equation}
    \mathcal{L}_\text{special} = 
    \lambda_\text{rep} \mathcal{L}_\text{rep} + 
    \lambda_\text{dsc} \mathcal{L}_\text{dsc} + 
    \lambda_\text{aux} \mathcal{L}_\text{aux}.
\end{equation}
\vspace{-0.8cm}
\subsection{Router-Only Task Adaptation for Expert Selection}
\label{subsec:adaptive_routing}
\vspace{-0.15cm}
A central challenge in task adaptation is to determine how to selectively access semantically meaningful components of the pretrained latent space based on task demands.
Unlike conventional supervised learning, which jointly updates both encoders and classifiers~\cite{10.1145/3219819.3220007, NEURIPS2022_b653f34d}, our approach takes a structurally constrained path: we freeze all pretrained experts and attention modules, and fine-tune only the lightweight router $g$, which accounts for a negligible fraction of the total parameters (\Cref{app:computational_cost}).
Rather than modifying the latent space, the router adaptively controls access to it, enabling task-specific selection over semantic components.

\vspace{-0.15cm}
This is nontrivial because task relevance is inherently context-dependent: even for the same input, the required semantics may differ across tasks.
Instead of explicitly defining task-relevant features, we leverage class labels as weak supervision, treating them as proxies for semantic consistency.
We assume that label-consistent samples are likely to express common latent factors essential for the task.
Accordingly, the router is trained to maximize mutual information among label-consistent samples, encouraging it to route inputs through experts that capture task-relevant semantics.
\vspace{-0.35cm}
\paragraph{Optimization Objective.}
To formalize the goal of task-driven routing, we define an objective that reflects the two principles introduced in \Cref{subsec:optimal_representation}. 
The aim is to construct routed representations $(Z_Y^1, Z_Y^2)$ that approximate the optimal task-specific representations $(Z_Y^{1*}, Z_Y^{2*})$, by encouraging them to retain task-relevant information while discarding task-irrelevant details from the input. Formally, we express this as the following objective, where $\alpha$ balances the two terms:
\vspace{-0.15cm}
\begin{equation}
    \max_{g} \boldsymbol{[}
        \underbrace{I(Z_Y^1, Z_Y^2; Y)}_\text{Task-Sufficiency}
        - \alpha \cdot \underbrace{I(Z_Y^1, Z_Y^2;X^1, X^2 | Y)}_\text{Information-Minimality}
    \boldsymbol{]}.
    \label{eq:selection_objective}
\end{equation}
\vspace{-0.7cm}
\paragraph{Training Loss Formulation.}
Direct computation of \cref{eq:selection_objective} is intractable, so we approximate each term using variational objectives that operationalize our design objective: the router should learn to activate experts that respond to task-relevant latent factors in the input, while suppressing task-irrelevant variations.
To approximate the Task-Sufficiency term, we adopt the supervised contrastive (SupCon) loss~\cite{khosla2020supervised}, originally introduced to enforce alignment between positive pairs defined by class labels.
In our framework, we reinterpret SupCon as a mechanism to maximize mutual information among label-consistent representations.
For a sample $x_i^m$ with positive index set $\mathcal{S}_{y_i} = \{s \mid y_s = y_i\}$, the SupCon loss is:
\vspace{-0.1cm}
\begin{equation}
    {\cal L}_{\text{SupCon}}^{[m \rightarrow {\bar m}]} = - {\mathbb E}_i {\mathbb E}_s \left[ \log \frac{\exp \left( \langle z_i^m, z_s^{\bar{m}} \rangle / \tau \right)}{\sum_j \exp \left( \langle z_i^m, z_j^{\bar{m}} \rangle/\tau \right)} \right].
\end{equation}
We show in \Cref{app:proof_supcon} that this loss provides a valid lower bound on task-conditioned mutual information. The final Task-Sufficiency loss averages over all four inter- and intra-modal directions:
\vspace{-0.15cm}
\begin{equation}
    {\cal L}_{\text{suff}} = \frac{1}{4} \sum_{m, {\bar m}}
    {\cal L}_{\text{SupCon}}^{[m \rightarrow \bar{m}]}.
\end{equation}
To estimate the Information-Minimality term, we consider its expression as a KL divergence $I(Z;X|Y) = {\mathbb E}_{p(x,y)}\left[ D_\text{KL}\left( p(z|x) || p(z|y) \right) \right]$.
We approximate both conditional distributions using von Mises-Fisher (vMF) distributions, leveraging the fact that, during the Specialization stage, encoder outputs lie on the unit hypersphere due to InfoNCE normalization.
Specifically, $p(z|x)$ is modeled as a vMF distribution centered at the instance representation $\mu_x = f(x)$, while $p(z|y)$ is modeled using the mean of label-consistent representations in the batch, $\hat{\mu}_y=\frac{\mu_y}{\parallel \mu_y\parallel}$, where $\mu_y= \frac{1}{|\mathcal{S}_y|} \sum_{s \in \mathcal{S}_y} z_s$.
Under the equal concentration assumption ($\kappa_x = \kappa_y$), the KL divergence reduces to an inner product form, yielding the following compactness loss:
\vspace{-0.1cm}
\begin{equation}
    {\mathcal L}_{\text{Comp}}^{[m \rightarrow \bar m]} = - \mathbb{E}_{p(x^m,y)}
    \left[ \langle {\mu_x^{m}}, \hat{\mu}_y^{\bar m} \rangle \right].
\end{equation}
Averaging across modality pairs gives the Information-Minimality loss:
\vspace{-0.15cm}
\begin{equation}
    {\cal L}_{\text{min}} = \frac{1}{4} \sum_{m, \bar{m}}
    {\cal L}_{\text{Comp}}^{[m \rightarrow \bar{m}]}.
\end{equation}
The final objective for the Selection stage is a weighted sum:
\vspace{-0.15cm}
\begin{equation}
{\cal L}_\text{select.} = \lambda_\text{suff}{\cal L}_\text{suff} + \lambda_\text{min} {\cal L}_\text{min}.
\end{equation}
Unlike Specialization, this stage does not include auxiliary routing losses, since the objective is not to balance expert usage but to selectively activate experts that are relevant to the target task. A complete derivation of the Information-Minimality estimator is provided in \Cref{app:proof_kl_vmf}.
\begin{figure*}[t]
    \centering
    \includegraphics[width=\textwidth]{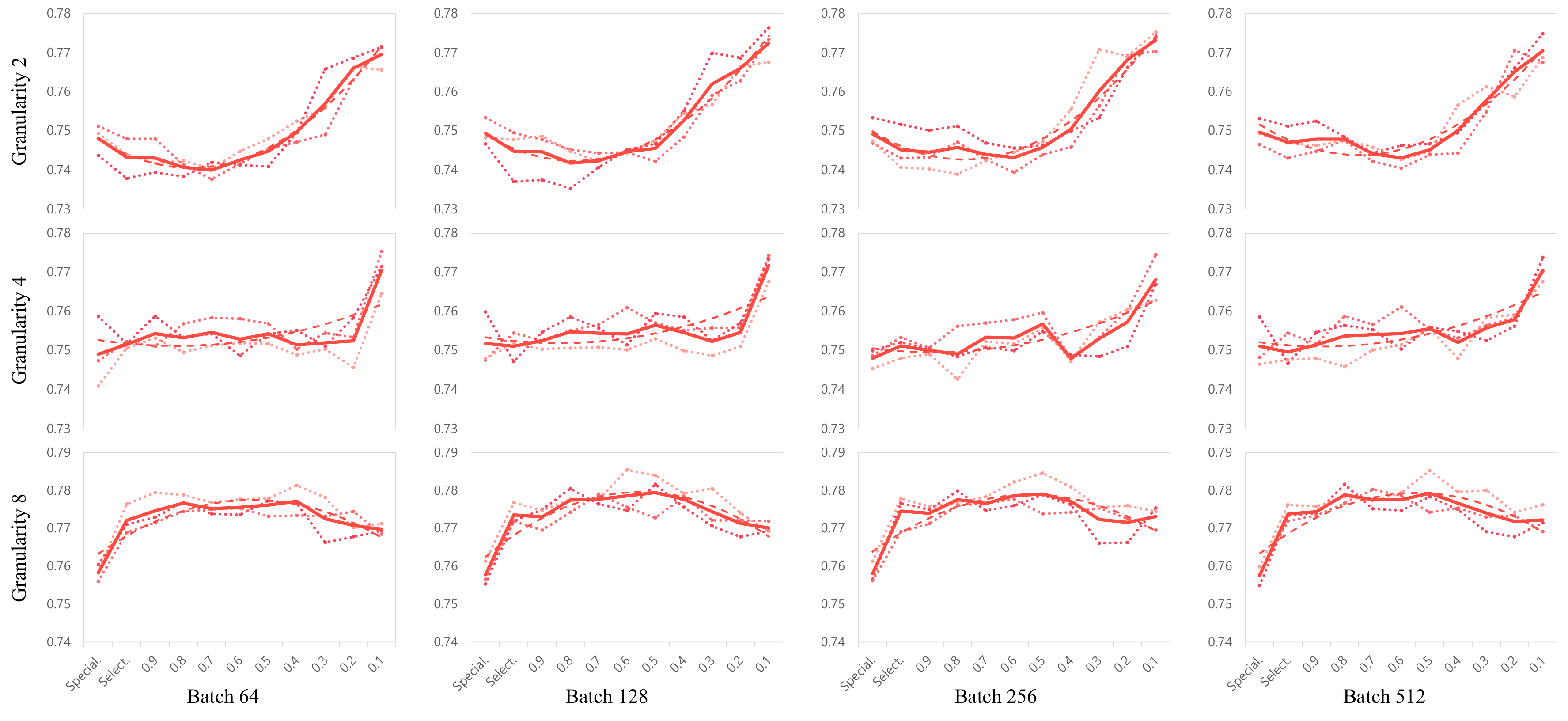}
    \vspace{-0.6cm}
    \caption{Performance on \textsc{MOSEI} across batch sizes (64-512) and $\chi$ (2,4,8). Dotted lines show individual random seeds, the dashed line their trend, and the solid line the mean. All results follow our three-stage pipeline, with $p$ decreased progressively during Sparsification. 
    }
    \vspace{-0.5cm}
    \label{fig:mosei_vis}
\end{figure*}
\vspace{-0.3cm}
\subsection{Inference-Time Pruning for Sparsification}
\label{subsec:routing_pruning}
\vspace{-0.1cm}
The final stage of the S3 framework, Sparsification, aims to make the representation more efficient at inference time.
In the previous Selection stage, the router was optimized to favor experts whose activations yield more task-relevant representations, implicitly assigning higher scores to those more informative for downstream prediction.
Consequently, the routing scores can be interpreted as quantitative indicators of each input-expert pair’s contribution to the task.
However, conventional MoE layers activate a fixed number of experts (e.g., top-$k$) per input, regardless of their actual utility. This can lead to redundant computation and allow irrelevant experts to contribute.

\vspace{-0.16cm}
To mitigate this, we introduce an inference-time pruning procedure that requires no additional training.
For a given batch, we collect the top-$k$ routing scores for all input-expert pairs and sort them. We then retain only the top $p \in [0,1]$ proportion of pairs, referred to as the \textit{preservation ratio}, and prune the remaining $(1-p)$ proportion from computation. Since residual connections remain active, removing a routing path does not completely sever the overall flow of information.
\vspace{-0.35cm}
\paragraph{Expected Behavior.}
Sparsification effectively regulates the amount of information retained in the representation by adjusting the preservation ratio $p$. As $p$ gradually decreases from 1 to 0, we expect downstream performance to follow a reverse U-shaped trend, the characteristic pruning curve:
\vspace{-0.3cm}
\begin{enumerate}
    \item \textbf{Task-Irrelevant Pruning.} At high $p$ values, pruning removes predominantly task-irrelevant paths, yielding a cleaner representation. As a result, performance typically improves or remains the same as when $p=1$.
    \vspace{-0.6cm}
    \item \textbf{Optimal Sparsity Regime (Sweet Spot).} At an appropriate sparsity level, unnecessary information is removed while task-relevant routes remain intact. The representation becomes minimal yet still sufficient, and performance reaches its peak.
    \vspace{-0.21cm}
    \item \textbf{Over-Pruning.} When $p$ becomes too small, pruning begins to remove task-relevant routes, discarding essential information and causing performance degradation.
\end{enumerate}
\vspace{-0.65cm}
\section{Experiments}
\label{sec:experiments}
\vspace{-0.1cm}

We evaluate our method on four multimodal benchmarks from the MultiBench~\cite{liang2021multibench}, following its standardized evaluation protocol: \textsc{MOSEI}~\cite{bagher-zadeh-etal-2018-multimodal}, \textsc{MOSI}~\cite{zadeh2016mosi}, \textsc{UR-FUNNY}~\cite{hasan-etal-2019-ur}, and \textsc{MUStARD}~\cite{castro-etal-2019-towards}. We compare S3 against representative baselines from three families:
(1) Contrastive learning (CLIP~\cite{radford2021learning});
(2) InfoMax-based approaches (FOCAL~\cite{liu2023focal}, DisentangledSSL~\cite{wang2025an}, JointOpt~\cite{pan2021disentangled});
(3) Augmentation-driven methods (FactorCL~\cite{liang2023factorized}).
Following \citet{wang2025an}, we report linear probing accuracy, averaged over three random seeds.

\vspace{-0.15cm}
Since MoE architectures inherently introduce a larger total parameter budget than dense models, we follow prior works~\cite{jiang2024mixtral, pmlr-v235-ludziejewski24a} and control for the \textit{active parameters}, those directly participating in token-wise transformations (attention and expert modules, excluding the router), for a fair comparison. 
Specifically, we set the top-$k$ value equal to the granularity ($k=\chi$) in each experiment, so that the number of activated expert parameters per token matches that of the original FFN.
\vspace{-0.25cm}
\subsection{Analysis of S3 Stages on MOSEI}
\label{subsec:analisys_of_S3_stages_on_MOSEI}
\vspace{-0.1cm}
We analyze how each stage of S3 influences downstream performance using \textsc{MOSEI}, the largest benchmark in our evaluation suite. As shown in \cref{fig:mosei_vis}, the pruning curves exhibit consistent trends across different batch sizes (64–512), while their shapes vary substantially with the chosen granularity. This suggests that the performance dynamics are governed more by the structural properties of the representation than by training configurations.

\vspace{-0.15cm}
At low granularity ($\chi=2$), the FFN is decomposed into a small number of large experts, causing multiple semantic concepts to be entangled within each expert. Such high degree of entanglement introduces routing ambiguity, making it difficult for the router to reliably identify task-relevant experts.
Consequently, performance degrades during Selection and early Sparsification, exhibiting a delayed U-shaped trend where performance recovers only after pruning removes enough noise that the remaining pathways carry a cleaner signal.

\vspace{-0.15cm}
In contrast, high granularity ($\chi=8$) divides the FFN into many small experts, yielding a finer expert-level separation of semantic concepts. In this setting, the router can more reliably assign high scores to task-relevant experts, allowing early pruning to effectively remove task-irrelevant paths. As a result, performance improves initially, peaks at an optimal sparsity level, and declines only when pruning begins to discard essential task-relevant routes, consistent with the characteristic reverse U-shaped pattern hypothesized in \cref{subsec:routing_pruning}. The intermediate setting ($\chi = 4$) exhibits behavior between these two extremes, with smoother performance transitions across pruning ratios.
\begin{table*}[t]
\setlength{\tabcolsep}{1.42pt}
\centering
\resizebox{\textwidth}{!}{
\renewcommand{\arraystretch}{1.3}
\begin{tabular}{L{2.0cm}C{0.6cm}C{1.9cm}C{1.9cm}C{1.9cm}C{1.9cm}C{1.9cm}C{1.9cm}C{1.9cm}C{1.9cm}C{1.9cm}C{1.9cm}C{1.9cm}}
\Xhline{3.5\arrayrulewidth}
\textbf{\textsc{Dataset}} & $\chi$ & Special. & Select. & 0.9 & 0.8 & 0.7 & 0.6 & 0.5 & 0.4 & 0.3 & 0.2 & 0.1 \\
\hline
\multirow{3}{*}{\textbf{\textsc{MOSEI}}} & 2 & \mosei{10.32}{74.94(0.35)} & \mosei{6.00}{74.48(0.68)} & \mosei{5.88}{74.46(0.62)} & \mosei{5.00}{74.18(0.56)} & \mosei{5.06}{74.24(0.18)} & \mosei{5.94}{74.47(0.04)} & \mosei{6.46}{74.55(0.29)} & \mosei{15.18}{75.27(0.37)} & \mosei{35.90}{76.20(0.70)} & \mosei{47.77}{76.60(0.29)} & \mosei{70.64}{77.25(0.45)} \\
& 4 & \mosei{13.72}{75.18(0.70)} & \mosei{12.65}{75.11(0.37)} & \mosei{14.84}{75.25(0.22)} & \mosei{18.98}{75.48(0.40)} & \mosei{18.21}{75.44(0.31)} & \mosei{17.84}{75.42(0.59)} & \mosei{22.44}{75.65(0.33)} & \mosei{18.40}{75.45(0.43)} & \mosei{14.52}{75.23(0.36)} & \mosei{18.59}{75.46(0.32)} & \mosei{67.97}{77.18(0.36)} \\
& 8 & \mosei{25.31}{75.78(0.32)} & \mosei{74.93}{77.36(0.29)} & \mosei{72.97}{77.31(0.30)} & \mosei{91.12}{77.75(0.31)} & \mosei{91.99}{77.77(0.13)} & \mosei{95.96}{77.86(0.60)} & \mosei{100.00}{\textcolor{white}{\textbf{77.95(0.59)}}} & \mosei{91.99}{77.77(0.19)} & \mosei{78.13}{77.44(0.53)} & \mosei{66.09}{77.13(0.32)} & \mosei{61.69}{77.01(0.16)} \\
\hline
\multirow{3}{*}{\textbf{\textsc{MOSI}}} & 2 & \mosi{7.76}{59.86(0.59)} & \mosi{7.33}{59.77(1.72)} & \mosi{8.80}{60.06(0.81)} & \mosi{5.74}{59.33(1.14)} & \mosi{5.00}{58.84(0.30)} & \mosi{8.26}{59.96(0.45)} & \mosi{9.94}{60.25(1.68)} & \mosi{19.65}{61.42(0.99)} & \mosi{27.94}{62.15(3.67)} & \mosi{21.11}{61.56(1.09)} & \mosi{71.51}{64.82(1.85)} \\
& 4 & \mosi{36.10}{62.76(0.81)} & \mosi{53.36}{63.85(2.41)} & \mosi{47.13}{63.48(1.35)} & \mosi{26.46}{62.03(1.26)} & \mosi{50.79}{63.70(0.55)} & \mosi{63.91}{64.43(2.10)} & \mosi{84.65}{65.45(1.90)} & \mosi{45.99}{63.41(1.24)} & \mosi{63.91}{64.43(2.24)} & \mosi{52.15}{63.78(1.74)} & \mosi{71.11}{64.80(2.27)} \\
& 8 & \mosi{48.44}{63.56(2.18)} & \mosi{61.28}{64.29(1.27)} & \mosi{62.96}{64.38(0.30)} & \mosi{78.47}{65.16(0.25)} & \mosi{100.00}{\textcolor{white}{\textbf{66.13(0.51)}}} & \mosi{86.83}{65.55(1.40)} & \mosi{87.93}{65.60(1.31)} & \mosi{78.47}{65.16(1.41)} & \mosi{87.93}{65.60(1.79)} & \mosi{96.74}{65.99(0.55)} & \mosi{75.36}{65.01(1.39)} \\
\hline
\multirow{3}{*}{\textbf{\textsc{UR-FUNNY}}} & 2 & \urfunny{15.24}{62.79(1.10)} & \urfunny{16.58}{62.85(1.00)} & \urfunny{5.51}{62.10(0.71)} & \urfunny{5.69}{62.13(0.80)} & \urfunny{5.00}{61.94(1.28)} & \urfunny{12.79}{62.67(1.37)} & \urfunny{17.51}{62.89(1.05)} & \urfunny{9.99}{62.51(0.99)} & \urfunny{9.99}{62.51(0.95)} & \urfunny{18.23}{62.92(1.67)} & \urfunny{77.42}{64.46(0.76)} \\
& 4 & \urfunny{24.92}{63.17(0.72)} & \urfunny{63.59}{64.18(0.99)} & \urfunny{71.33}{64.34(1.42)} & \urfunny{56.27}{64.02(0.72)} & \urfunny{80.56}{64.52(0.58)} & \urfunny{78.98}{64.49(0.60)} & \urfunny{60.33}{64.11(0.86)} & \urfunny{92.55}{64.74(0.74)} & \urfunny{60.33}{64.11(0.61)} & \urfunny{57.61}{64.05(0.47)} & \urfunny{17.51}{62.89(0.80)} \\
& 8 & \urfunny{39.53}{63.61(0.49)} & \urfunny{36.26}{63.52(0.62)} & \urfunny{38.43}{63.58(0.66)} & \urfunny{37.33}{63.55(0.36)} & \urfunny{34.85}{63.48(0.24)} & \urfunny{52.35}{63.93(0.77)} & \urfunny{36.26}{63.52(0.19)} & \urfunny{74.35}{64.40(0.72)} & \urfunny{89.20}{64.68(0.96)} & \urfunny{100.00}{\textcolor{white}{\textbf{64.87(0.63)}}} & \urfunny{34.85}{63.48(1.13)} \\
\hline
\multirow{3}{*}{\textbf{\textsc{MUStARD}}} & 2 & \mustard{28.64}{57.49(3.57)} & \mustard{28.64}{57.49(3.02)} & \mustard{38.85}{58.45(1.51)} & \mustard{18.15}{56.28(2.93)} & \mustard{60.67}{60.14(6.91)} & \mustard{41.78}{58.70(6.91)} & \mustard{44.68}{58.94(6.03)} & \mustard{31.04}{57.73(4.66)} & \mustard{67.88}{60.63(5.86)} & \mustard{60.67}{60.14(0.00)} & \mustard{79.17}{61.35(3.99)} \\
& 4 & \mustard{21.99}{56.76(3.57)} & \mustard{31.04}{57.73(3.42)} & \mustard{24.07}{57.00(2.54)} & \mustard{79.17}{61.35(1.51)} & \mustard{38.85}{58.45(2.93)} & \mustard{38.85}{58.45(3.35)} & \mustard{44.68}{58.94(2.33)} & \mustard{33.55}{57.97(5.02)} & \mustard{20.01}{56.52(5.07)} & \mustard{5.00}{53.14(2.93)} & \mustard{5.00}{53.14(5.81)} \\
& 8 & \mustard{41.78}{58.70(3.16)} & \mustard{83.12}{61.59(0.72)} & \mustard{87.33}{61.84(0.84)} & \mustard{100.00}{\textcolor{white}{\textbf{62.56(1.11)}}} & \mustard{87.33}{61.84(1.11)} & \mustard{71.55}{60.87(0.72)} & \mustard{75.32}{61.11(1.11)} & \mustard{67.88}{60.63(1.51)} & \mustard{83.12}{61.59(1.45)} & \mustard{38.85}{58.45(0.84)} & \mustard{47.69}{59.18(4.82)} \\
\Xhline{3.5\arrayrulewidth}
\end{tabular}
}
\vspace{0.08cm}
\caption{Prediction accuracy (\%) of the S3 framework across benchmarks and granularities, covering Specialization, Selection, and Sparsification (evaluated over preservation ratios). All results are averaged over three random seeds with standard deviations.}
\vspace{-0.7cm}
\label{tab:benchmark-wide_evaluation}
\end{table*}
\vspace{-0.25cm}
\subsection{Benchmark-Wide Evaluation of the S3 Framework}
\vspace{-0.1cm}
\Cref{tab:benchmark-wide_evaluation} summarizes the performance progression of S3 across the four multimodal benchmarks. Despite differences in task characteristics and dataset scales, the structural pruning patterns analyzed in \cref{subsec:analisys_of_S3_stages_on_MOSEI} consistently recur across all benchmarks.

\vspace{-0.15cm}
With sufficiently high granularity, all benchmarks exhibit a monotonic performance improvement from Specialization to Selection, reaching peak performance during Sparsification. 
This consistency demonstrates that the benefits of fine-grained semantic decomposition and router-based information refinement are not benchmark-specific artifacts, but rather stem from the intrinsic inductive advantages of the S3 architecture. 
Conversely, low-granularity settings exhibit similar degradation during Selection and early instability in Sparsification, indicating that coarse-grained decomposition leads to less stable performance dynamics.

\vspace{-0.15cm}
Beyond these architectural effects, we also observe that the scale of the dataset strongly influences training stability. Larger datasets such as \textsc{MOSEI} and \textsc{UR-FUNNY} exhibit minimal variance across random seeds, whereas variance is substantially higher for smaller datasets such as \textsc{MOSI} and \textsc{MUStARD}.
This effect is most notable at $\chi=2$, where limited data and coarse decomposition jointly lead to more variable routing behavior.
\vspace{-0.2cm}
\subsection{Comparison with Prior Methods}
\label{subsec:comparison_with_prior_methods}
\vspace{-0.1cm}
As shown in \cref{tab:comparison_with_prior_works}, the S3 framework consistently outperforms a broad range of prior methods, spanning contrastive learning, InfoMax-based, and augmentation-driven approaches. DisentangledSSL, which builds on an information-theoretic analysis of MMRL via the InfoMax principle, performs competitively across all benchmarks. 
However, it is consistently outperformed by S3, highlighting the benefits of our structured approach. These gains stem from S3’s ability to satisfy the two criteria at the core of our formulation, Task-Sufficiency and Information-Minimality.

\vspace{-0.1cm}
The Specialization stage aims to retain a rich set of semantic factors. However, our results show that preserving more information alone does not necessarily guarantee improved downstream performance. In contrast, the Selection stage employs a router to activate task-relevant semantic experts, addressing both the shared-only limitation of contrastive learning and the indiscriminate retention of InfoMax methods.
The most significant improvements arise in the Sparsification stage, where pruning low-contribution pathways sharpens the representation without modifying the underlying encoders. This refinement is not merely 
for computational efficiency; but introduces a finer-grained, sample-level pruning that complements the task-level routing of the Selection stage, providing a structural advantage unique to S3 and absent from prior methods.

\vspace{-0.15cm}
Similarly, augmentation-based methods aim to capture task-relevant features by assuming that they remain invariant across augmented views, while allowing task-irrelevant factors to vary.
However, their effectiveness is highly sensitive to the choice of augmentation strategies and often fluctuates across benchmarks. 
In contrast, S3 achieves stable and reproducible gains without relying on such heuristics, instead using expert routing to structurally disentangle latent semantic factors and selectively recombine them.

\begin{table}[t]
\setlength{\tabcolsep}{1.45pt}
\begin{center}
\resizebox{\linewidth}{!}{
\renewcommand{\arraystretch}{1.2}
\begin{tabular}{L{2.5cm}C{2.0cm}C{2.0cm}C{2.0cm}C{2.0cm}}
\Xhline{3.5\arrayrulewidth}
 & \textbf{\textsc{MOSEI}} & \textbf{\textsc{MOSI}} & \textbf{\textsc{UR-FUNNY}} & \textbf{\textsc{MUStARD}} \\
\hline
CLIP & 76.87(0.45) & 64.24(0.88) & 62.73(0.92) &  56.04(4.19) \\
FactorCL-emb & 71.80(0.64) & 62.97(0.81) & 63.29(2.07) & 56.76(4.66) \\
FactorCL-proj & 74.61(1.65) & 56.02(1.26) & 61.25(0.47) & 55.80(2.18) \\
FOCAL & 76.77(0.51) & 63.65(1.09) & 63.17(0.96) & 58.21(2.21) \\
JointOpt & 76.71(0.14) & 65.02(1.96) & 63.58(1.45) & 57.73(4.12) \\
DisentangledSSL & 77.45(0.06) & 65.16(0.81) & 64.24(1.54) & 61.60(2.61) \\
\hline
Specialization & 75.78(0.32) & 63.56(2.18) & 63.61(0.49) & 58.70(3.16) \\
+ Selection & 77.36(0.29) & 64.29(1.27) & 63.52(0.62) & 61.59(0.72) \\
+ Sparsification & \textbf{77.95(0.95)} & \textbf{66.13(0.51)} & \textbf{64.87(0.63)} & \textbf{62.56(1.11)}\\
\Xhline{3.5\arrayrulewidth}
\end{tabular}
}

\end{center}
\vspace{-0.05cm}
\caption{Prediction accuracy (\%) on four benchmarks, averaged over three random seeds, with standard deviations. We compare S3 against representative prior methods.}
\vspace{-0.95cm}
\label{tab:comparison_with_prior_works}
\end{table}
\vspace{-0.3cm}
\section{Conclusion}
\label{sec:conclusion}
\vspace{-0.1cm}
This work advocates a structural perspective on MMRL. Rather than compressing heterogeneous signals into monolithic embeddings, we emphasize the need to decompose inputs into semantically coherent components, selectively retain task-relevant information, and discard redundant variability. We realize this principle through S3, a three-stage MoE-based framework that progressively specializes, selects, and sparsifies multimodal representations.
Empirically, S3 consistently improves performance across diverse benchmarks while exhibiting stable and predictable behavior across semantic granularities and dataset scales. These results suggest that S3’s effectiveness is not merely a result of objective tuning, but of structural inductive biases aligned with the compositional nature of multimodal semantics.

\vspace{-0.1cm}
Looking ahead, this structural paradigm opens several promising research directions:
(1) modality-adaptive information preservation for task-critical modalities;
(2) layer-adaptive modeling of semantic abstraction across model depth;
(3) self-supervised routing adaptation to reduce label dependence;
(4) enhanced expert specialization for more precise semantic decomposition; and
(5) adaptive sparsification with learnable granularity and pruning policies.
\section*{Acknowledgements}
This work was supported by the Korean Government through the grants from IITP (RS-2021-II211343, RS-2022-II220953, RS-2025-25442338).

\section*{Impact Statement}
This paper presents work whose goal is to advance the field of Machine Learning. There are many potential societal consequences of our work, none which we feel must be specifically highlighted here.


\bibliography{main}
\bibliographystyle{icml2026}

\newpage
\appendix
\onecolumn
\clearpage

\section{Conceptual Overview of Our Framework}
\label{app:conceptual}

\begin{wrapfigure}{r}{0.5\linewidth}
    \vspace{-0.7\baselineskip}
    \centering
    \includegraphics[width=\linewidth]{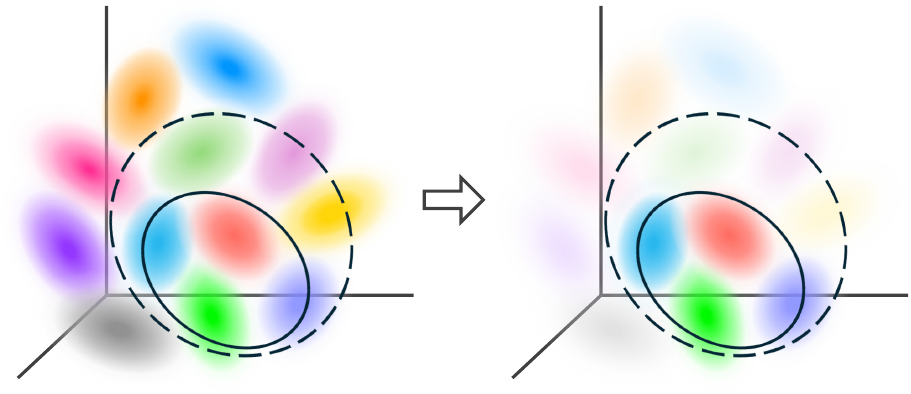}
    \caption{Multimodal observations contain heterogeneous semantic components. Our framework decomposes these components (in different colors) and maps them into a modality-agnostic space, preserving only the task-relevant subset. 
    Dashed ellipses denote the multimodal joint information, and solid ellipses indicate the task-relevant region.
    }
    \label{fig:conceptual_sketch}
    \vspace{-0.3cm}
\end{wrapfigure}

Multimodal data inherently contain heterogeneous semantic components, not all of which are equally relevant to the target task. 
Our core insight is that these semantic components can be decomposed, selectively represented, and filtered in the representation space to achieve Task-Sufficient and Information-Minimal representations.
As illustrated in Figure~\ref{fig:conceptual_sketch}, our framework first disentangles the multimodal observation into its underlying semantic constituents.
We then map these into a modality-agnostic latent space, where only the components deemed relevant to the task are selectively preserved and activated.
This figure provides the conceptual foundation for our method, where the goal is not merely to fuse modalities, but to structurally filter and retain only the minimal sufficient set of features needed for the task at hand.
This idea serves as the design principle for our approach S3, motivating our decomposition of multimodal features, expert specialization, and task-driven expert selection via routing.

\section{Conceptual Explanation of Distribitional Semantic Coherence}
\label{app:distribitional_semantic_coherence}
We provide a conceptual explanation of Distributional Semantic Coherence (DSC), as formally defined in Definition~\ref{def:distributional_semantic_coherence_formal}. The central idea of DSC is that multimodal semantic alignment should be understood at the level of latent semantic concepts, rather than through sample-wise or instance-level correspondence.

Multimodal data are inherently heterogeneous, and the information observed by different modalities often overlaps only partially and asymmetrically. Moreover, the set of semantic concepts shared across modalities can vary over time and context. For instance, in a video containing both a drum and a person, some segments may involve the drum being played (visual: drum and person, audio: drum sound), while others may involve the person speaking (visual: drum and person, audio: human voice). In such cases, the semantic correspondence between modalities shifts across segments, making rigid instance-level alignment assumptions inadequate. As a result, enforcing sample-level hard alignment, such as directly matching drum sounds to drum+person instances, fails to adequately capture the contextual and asymmetric nature of multimodal semantics.

To address this challenge, our model processes each modality using a Mixture-of-Experts (MoE) architecture,
where each expert is encouraged to specialize in a distinct latent semantic concept. Inputs are routed probabilistically rather than deterministically, allowing each feature to contribute to multiple concepts. Importantly, features from different modalities that correspond to the same semantic concept can co-activate the same expert or a consistent subset of experts. This design enables semantic concepts to serve as shared intermediates across modalities, rather than enforcing direct alignment between raw modality-specific features.

Under this formulation, cross-modal alignment does not arise from enforcing direct correspondence between individual samples. Instead, alignment emerges through repeated co-activation of concept-specific experts associated with the same semantic concept across modalities. In some samples, a drum-related expert may be jointly activated by both audio and visual inputs, while in others, a person-related expert may dominate the alignment signal. Across diverse contexts and combinations of multimodal inputs, each semantic concept is repeatedly routed through the same expert(s) and updated via cross-modal contrastive learning. Through this process, semantic alignment is accumulated at the level of the data distribution, rather than being imposed at the level of individual samples.

The term distributional emphasizes that semantic coherence is established through statistical regularities observed across the dataset, not through explicit, one-to-one alignments within single samples. Each semantic concept appears in diverse contexts and alongside varying co-occurring entities. By consistently activating the same expert(s) across such variations, the model learns modality-invariant representations that remain robust to contextual changes. We refer to this property as Distributional Semantic Coherence, reflecting a form of multimodal alignment that respects the heterogeneous and context-dependent nature of real-world multimodal data.

\section{Related Works}
\label{sec:related_works}
\subsection{Multimodal Representation Learning}
Multimodal representation learning (MMRL) aims to integrate information from heterogeneous modalities into semantically coherent representations. A dominant paradigm in this field is multimodal contrastive learning, which aligns representations from different modalities within a shared embedding space~\cite{khan2025survey, Choi_2025_ICCV, jiang2023understanding, Park_2024_WACV}. CLIP-style models exemplify this approach, achieving strong performance across diverse downstream tasks~\cite{radford2021learning,li2021align,li2022blip,cherti2023reproducible}.
However, in realistic scenarios where modalities exhibit partial overlap or asymmetry, contrastive objectives can become sensitive to modality-specific signals, leading to unstable or distorted alignments~\cite{jiang2023understanding,liang2022mind,schrodi2025two,liao2025multimodal}. To address this, several studies introduce regularization to suppress unique modality information~\cite{almudevar2025aligning}. However, such shared-information centric alignment may discard task-relevant signals when downstream tasks rely on modality-unique cues, a limitation that has been both theoretically and empirically observed~\cite{liang2023factorized,wang2025an}.

An alternative line of research adopts InfoMax-based objectives, aiming to preserve as much information as possible within each modality~\cite{jiang2023understanding,dong2023simmmdg}. These approaches typically employ latent factorization or disentangled representation structures to explicitly model both shared and modality-unique components. For instance, \citet{wang2025an} propose DisentangledSSL, which sequentially learns shared and modality-unique representations using an information-theoretic criterion. Similarly, \citet{liu2023focal} introduce a contrastive framework that orthogonally decomposes multimodal time-series data into shared and unique latent features, while jointly enforcing modality consistency, transformation consistency, and temporal locality. Related ideas have also been explored in domains such as single-cell multimodal analysis and T-cell receptor design~\cite{Zhang2024.10.01.615977,li2023disentangled}.

Nevertheless, InfoMax-oriented approaches tend to preserve task-irrelevant variability alongside useful information, which can adversely affect downstream performance~\cite{49568,soattoC16}. To address this issue, \citet{liang2023factorized} propose an objective based on multi-view redundancy~\cite{tian2020contrastive,bachman2019learning,hjelm2018learning}, estimating task-relevant information through multimodal augmentation. By enforcing invariance across augmented views and leveraging bounds on mutual information, their method aims to suppress unnecessary variability. However, this approach relies on the assumption that task-relevant factors remain invariant under the chosen augmentations, making performance sensitive to augmentation quality and design.

In contrast to refining objectives alone, this work focuses on the structural organization of multimodal representations. 
We consider representations that decompose inputs into semantically interpretable components and allow selective recombination based on downstream task requirements. Such structural flexibility enables a more fine-grained correspondence between information and task demands, providing a complementary perspective to existing MMRL paradigms.

\subsection{Mixture-of-Experts Models}
Mixture-of-Experts (MoE) models employ conditional computation by selectively activating a subset of experts based on the input, allowing for efficient scaling of model capacity without a proportional increase in computation~\cite{shazeer2017,lepikhin2021gshard}.
This design has been adopted in Transformer-based architectures to support large-scale training under limited resources~\cite{rajbhandari2022deepspeed,fedus2022switch,du2022glam}.
In these models, experts are treated as computational units, and routing is primarily used to distribute computational load and maintain training stability. Consequently, expert selection is driven more by efficiency considerations than by semantic structuring of representations.
\paragraph{Multimodal Mixture-of-Experts.}
To address the increasing computational demands and scalability challenges in multimodal settings, recent works have adapted MoE architectures to support multimodal inputs. For example, \citet{mustafa2022multimodal}, \citet{shen2023scaling}, and \citet{li2025uni} apply sparse MoE to image-text and omnimodal models, activating only a subset of experts to enhance efficiency and scalability. These works largely build upon conventional MoE research, adapting it for large-scale multimodal model training. 
Beyond scaling, some approaches utilize MoE to enhance robustness under missing modality scenarios. \citet{NEURIPS2024_b2f2af54} and \citet{han2024fusemoe} propose routing mechanisms that dynamically select experts based on observed modalities or infer missing information. Other studies apply multimodal MoE architectures to domain-specific tasks such as continual learning~\cite{huai2025cl} and recommender systems~\cite{10.1145/3583780.3614978}.

While these methods focus on scaling, robustness, or domain adaptation, they typically lack explicit mechanisms for controlling the semantic structure of learned representations. Routing is primarily optimized for computational or adaptive efficiency, with limited emphasis on selectively preserving, suppressing, or structurally disentangling information based on task-specific semantics.
\paragraph{MoE for Interpretability and Semantic Decomposition.}
More recently, a new line of research explores MoE not merely as a tool for efficiency, but as a structural prior for semantic decomposition and model interpretability. These studies reinterpret experts as semantically meaningful units, rather than as purely computational blocks.
For instance, \citet{park2025monet} decomposes the feedforward layers of large language models into 262K monosemantic experts, each specialized for distinct semantic features. This approach mitigates the polysemantic neuron problem and enables fine-grained analysis of model activations, demonstrating both mechanistic interpretability and functional manipulability. Similarly, \citet{yang2025mixture} promotes expert-level semantic specialization by combining sparsity-aware routing with ReLU-induced activation sparsity. 
\citet{oldfield2025towards} critiques neuron-level sparse autoencoder approaches~\cite{huben2024sparse,bricken2023monosemanticity} for their performance degradation, and proposes the Mixture of Decoders, which introduces layer-level instead. 

While these approaches highlight the inductive potential of MoE for structured representation, their focus remains on analyzing model internals or enhancing interpretability. In contrast, they do not offer explicit mechanisms for regulating which semantic information should be retained or discarded based on downstream task requirements, nor do they address task-driven selection and pruning from an information-theoretic standpoint.
\paragraph{Multitask Mixture-of-Experts.}
MoE has also been widely adopted in multitask learning to mitigate task interference and enable flexible parameter sharing~\cite{Chen_2023_ICCV,NEURIPS2022_b653f34d}. For example, \citet{10.1145/3219819.3220007} propose a multi-gate MoE to address performance degradation in shared-bottom models when task relatedness is low, allowing each task to dynamically select its own expert composition. \citet{NEURIPS2022_b653f34d} dynamically adjust the number of active experts based on task difficulty to reduce gradient conflicts. Other works explore task-specific expert routing via memory-based selection~\cite{Ye_2023_ICCV} or domain-aware routing in multi-domain multitask settings~\cite{10.1145/3626772.3657686}.
These models are trained end-to-end with supervised losses, where routing serves to separate tasks and reduce interference, rather than to explicitly regulate the semantic content of representations.

In contrast, our approach leverages MoE as a structural prior for semantic decomposition. During self-supervised pretraining, experts are encouraged to specialize according to latent semantic factors. At fine-tuning, we freeze both the backbone and experts, adapting only the router to selectively activate task-relevant experts. Unlike conventional multitask MoE approaches, where routing is optimized solely for predictive accuracy, our router functions as an information-theoretic bottleneck: it is trained to control the semantic flow, preserving only task-sufficient information while discarding irrelevant components.

\section{Proofs of Problem Formulation}
\label{app:problem_formulation_proofs}
\subsection{Proof of Proposition~\ref{prop:dpi_sufficiency} (Data Processing Inequality)}
\label{app:proof_dpi_sufficiency}
For convenience, we denote $X = (X^1, X^2)$ and $Z = (Z^1, Z^2)$. Under the Markov chain assumption $Y \to X \to Z$ in \cref{eq:multimodal_markov_chain},  
the joint distribution factorizes as $p(y,x,z) = p(y)\,p(x|y)\,p(z|x)$,  
which implies that $Y$ and $Z$ are conditionally independent given $X$:
\begin{equation}
    p(y, z|x) = p(y|x)\, p(z|x) \quad \Longleftrightarrow \quad I(Z;Y|X)=0.
    \label{eq:conditional_independent}
\end{equation}
We now apply the chain rule of mutual information:
\begin{alignat}{2}
    I(X;Y)
    &= I(Z,X;Y) - I(Z;Y|X) &\qquad& \\
    &= I(Z,X;Y)
    &\qquad& (\because I(Z;Y|X)=0, \text{ \cref{eq:conditional_independent}})\\
    &= I(Z;Y) + I(X;Y|Z)
    &\qquad& (\text{by chain rule: } I(Z,X;Y) = I(Z;Y) + I(X;Y|Z)) \\
    &\ge I(Z;Y).
    &\qquad& (\because \text{non-negativity of conditional MI})
\end{alignat}
The equality holds if and only if $I(X;Y|Z) = 0$, which means that $Z$ is a sufficient statistic for $Y$ with respect to $X$. In this case, $(Z^1, Z^2)$ retains all task-relevant information in $(X^1, X^2)$ for predicting $Y$. \hfill $\square$

\subsection{Proof of Proposition~\ref{prop:mi_decomposition} (Mutual Information Decomposition)}
\label{app:proof_mi_decomposition}
According to Definition~\ref{def:latent_factor_model}, we decompose the input modalities as $X^1 = (X_S, X_U^1)$ and $X^2 = (X_S, X_U^2)$.  
Applying the chain rule of mutual information, we have:
\begin{align}
    I(X^1;X^2) &= I(X_S,X_U^1;X_S,X_U^2) \\
    &= I(X_S;X_S,X_U^2) + I(X_U^1; X_S, X_U^2|X_S)\\
    &= \big[ \, I(X_S;X_S) + I(X_S;X_U^2|X_S) \,\big] + \big[ \, I(X_U^1;X_U^2|X_S) + I(X_U^1;X_S|X_S,X_U^2) \, \big] \\
    &= I(X_S;X_S) + 0 + 0 + 0 \qquad \qquad (\because X_S \perp X_U^1, \; X_S \perp X_U^2, \; X_U^1 \perp X_U^2)\\
    &= H(X_S).
\end{align}
Therefore, the modality-unique factors $X_U^1$ and $X_U^2$ do not contribute to the mutual information between $X^1$ and $X^2$,  
which is solely determined by the shared factor $X_S$.  
\hfill $\square$

\subsection{Proof of Proposition~\ref{prop:cl_limitation} (Fundamental Limitation of Contrastive Representations)}
\label{app:proof_cl_limitation}
Proposition~\ref{prop:mi_decomposition} shows that the cross-modal mutual information $I(X^1; X^2)$ is entirely determined by the shared factor $X_S$. This implies that multimodal contrastive learning, by design, optimizes only for aligning shared information across modalities, without any pressure to preserve modality-unique factors $(X_U^1, X_U^2)$. As a result, the optimal solution induced by contrastive objectives corresponds to a shared-only representation that satisfies:
\begin{equation}
    (X_U^1, X_U^2) \perp (Z^1_\text{CL}, Z^2_\text{CL}) \mid X_S .
    \label{eq:cl_shared_only}
\end{equation}
Now consider a downstream task that relies on the unique factors. According to Definition~\ref{def:task_unique}, this implies:
\begin{equation}
    I(X_U^1, X_U^2;Y) > 0.
\end{equation}
In other words, the presence of nonzero mutual information $I(X_U^1, X_U^2; Y)$ indicates that $X_S$ alone is insufficient for predicting $Y$, as the unique factors $X_U^1$ and $X_U^2$ carry task-relevant information.
\begin{align}
    I(X^1, X^2;Y|Z^1_\text{CL}, Z^2_\text{CL}) 
    &= I(X_S, X_U^1, X_U^2; Y|Z^1_\text{CL}, Z^2_\text{CL}) \\
    &\geq I(X_U^1, X_U^2; Y|Z^1_\text{CL}, Z^2_\text{CL}) \\
    &\geq I(X_U^1, X_U^2; Y|X_S)  \qquad (\because \text{Markov chain \Cref{eq:cl_markov}})\\
    &= I(X_U^1, X_U^2; Y) \qquad \qquad (\because X_S \perp X_U^1, \; X_S \perp X_U^2) \\
    &> 0.
\end{align}
According to \Cref{app:proof_dpi_sufficiency}, if $I(X^1, X^2; Y | Z^1_\text{CL}, Z^2_\text{CL}) > 0$,  
then the representation $(Z^1_\text{CL}, Z^2_\text{CL})$ cannot be a sufficient statistic for $Y$.  
In this case, the strict inequality form of the data processing inequality (DPI) holds:
\begin{equation}
    I(X^1,X^2;Y) > I(Z^1_\text{CL}, Z^2_\text{CL};Y).
\end{equation}
\hfill $\square$

\section{Estimating Mutual Information Objectives}
\label{app:mi_estimation}
\subsection{InfoNCE as a Lower Bound on Mutual Information}
\label{app:proof_infonce}
The InfoNCE (Noise-Contrastive Estimation) loss is widely used as a variational lower bound estimator of mutual information $I(X; Z)$. The derivation presented here follows the standard formulation from \citet{journals/corr/abs-1807-03748}. The mutual information is defined as:
\begin{equation}
I(X; Z) = \mathbb{E}_{p(x, z)} \left[ \log \frac{p(x|z)}{p(x)} \right].
\label{eq:mi}
\end{equation}
In practice, the true densities $p(x)$ and $p(x|z)$ are not accessible, so we introduce a non-negative score function $\psi(x, z)$ to approximate the log-density ratio $\log \frac{p(x|z)}{p(x)}$. A common instantiation defines $\psi(x, z) := \langle f(x), z \rangle / \tau$, where $f(x)$ is an encoder, $\langle \cdot, \cdot \rangle$ denotes cosine similarity between $\ell_2$-normalized vectors, and $\tau$ is a temperature hyperparameter. Assuming that $\psi(x, z)$ approximates the log-density ratio, we obtain:
\begin{equation}
    \exp(\psi(x, z)) \propto \frac{p(x|z)}{p(x)} .
\end{equation}
Given a batch $\mathcal{B} = \{1, \ldots, B\}$, the InfoNCE loss for a positive pair $(x_i, z_i)$ is defined in the form of a cross-entropy:
\begin{equation}
\mathcal{L}_{\text{InfoNCE}} = - \mathbb{E}_{i} \left[ \log \frac{\exp(\psi(x_i, z_i))}{\sum_{j}^B \exp(\psi(x_i, z_j))} \right].
\label{eq:mi_appendix}
\end{equation}
Substituting the density ratio approximation into \cref{eq:mi_appendix}, we have:
\begin{align}
    {\cal L}_\text{InfoNCE} &= -{\mathbb E}_{i} \left[ \log \frac{\frac{p(x_i|z_i)}{p(x_i)}}{\frac{p(x_i|z_i)}{p(x_i)} + \sum_{j \ne i}^B \frac{p(x_i|z_j)}{p(x_i)}} \right] \\
    &= \mathbb{E}_{i} \left[ \log \left(1 + \frac{p(x_i)}{p(x_i|z_i)} \sum_{j \neq i}^B \frac{p(x_i|z_j)}{p(x_i)} \right) \right] \\
    &\approx \mathbb{E}_{i} \left[ \log \left(1 + \frac{p(x_i)}{p(x_i|z_i)}(B - 1) \ \mathbb{E}_{j} \left[ \frac{p(x_i|z_j)}{p(x_i)} \right]\right) \right]\\
    &= \mathbb{E}_{i} \left[ \log \left(1 + \frac{p(x_i)}{p(x_i|z_i)}(B-1) \right)\right] \\
    &\geq \mathbb{E}_{i}\left[ \log \left( \frac{p(x_i)}{p(x_i|z_i)}B \right)\right] \\
    &= - I(X;Z) + \log B.
\end{align}
Here, we make a standard assumption that the negative samples $z_j$ (for $j \ne i$) are drawn independently of the anchor $x_i$, implying $p(x_i|z_j) \approx p(x_i)$. Therefore, InfoNCE provides the following variational lower bound on mutual information:
\begin{equation}
I(X; Z) \geq \log B - \mathcal{L}_{\text{InfoNCE}}.
\label{eq:proof_infonce}
\end{equation}
Minimizing $\mathcal{L}_\text{InfoNCE}$ is thus equivalent to maximizing this lower bound, encouraging the encoder to preserve as much information about $X$ in $Z$ as possible.
\hfill $\square$

\subsection{SupCon as a Lower Bound on Task-Conditioned Mutual Information}
\label{app:proof_supcon}
In this section, we show that the supervised contrastive (SupCon) loss~\cite{khosla2020supervised} serves as a variational lower bound estimator of the mutual information $I(Z;Y)$ between the learned representation $Z$ and the label $Y$. The target mutual information is defined as:
\begin{equation}
    I(Z;Y) = {\mathbb E}_i \left[ \log \frac{p(z_i|y_i)}{p(z_i)} \right].
    \label{}
\end{equation}
Given a batch $\mathcal{B} = \{1, \ldots, B\}$ and a set of positive indices $\mathcal{S}_{y_i} = \{s \mid y_s = y_i\}$ that share the same label as anchor $i$, the SupCon loss is defined as:
\begin{equation}
    {\cal L}_\text{SupCon} = - {\mathbb E}_i {\mathbb E}_{s}\left[ \log \frac{\exp(\psi(z_i, z_s))}{\sum_j^B \exp(\psi(z_i, z_j))} \right].
    \label{eq:supcon_definition}
\end{equation}
where the score function is defined as $\psi(z_i, z_j) := \langle z_i, z_j \rangle / \tau$, using cosine similarity between $\ell_2$-normalized representations and a temperature parameter $\tau$. The index $s$ is uniformly sampled from the empirical positive set $\mathcal{S}_{y_i}$, and we write $\mathbb{E}_s$ to denote expectation over this set.
Since the log function is concave, we apply Jensen’s inequality to obtain a lower bound on the inner expectation:
\begin{align}
    {\mathbb E}_{s} \left[ \log \frac{\exp(\psi(z_i, z_s))}{ \sum_j^B \exp(\psi(z_i, z_j))} \right] &\leq \log \left( {\mathbb E}_{s} \left[ \frac{\exp(\psi(z_i, z_s))}{ \sum_j^B \exp(\psi(z_i, z_j))} \right] \right) \\
    &= \log \left( \frac{1}{|{\cal S}_{y_i}|} \sum_{s \in {\cal S}_{y_i}} \frac{\exp(\psi(z_i, z_s))}{ \sum_j^B \exp(\psi(z_i, z_j))} \right) \\ 
    &= \log \left(\frac{\frac{1}{|{\cal S}_{y_i}|}\sum_{s \in {\cal S}_{y_i}}\exp(\psi(z_i, z_s))}{\sum_j^B \exp(\psi(z_i, z_j))} \right).
\end{align}
Substituting this back into the definition of SupCon in \cref{eq:supcon_definition}, we obtain the following lower bound:
\begin{equation}
    {\cal L}_\text{SupCon} \geq - {\mathbb E}_i \left[ 
        \log \left(\frac{\frac{1}{|{\cal S}_{y_i}|}\sum_{s \in {\cal S}_{y_i}}\exp(\psi(z_i, z_s))}{\sum_j^B \exp(\psi(z_i, z_j))} \right).
    \right]
    \label{eq:supcon_jensen}
\end{equation}
We now reinterpret the exponential score $\exp(\psi(z_i, z))$ as a non-parametric kernel function in a kernel density estimation (KDE) framework, similar to the analysis in \cref{app:proof_infonce}. Under this view, the numerator in \cref{eq:supcon_jensen} corresponds to a batch-wise KDE estimator of the class-conditional density $p(z_i | y_i)$, constructed from the empirical positive set $\mathcal{S}_{y_i}$. The denominator serves as a KDE estimator of the marginal density $p(z_i)$ over the full batch. Accordingly, we define the estimators as:
\begin{equation}
    \hat{p}(z_i|y_i) =  \frac{1}{|{\cal S}_{y_i}|} \sum_{s \in {\cal S}_{y_i}} \exp(\psi(z_i, z_s)), 
    \qquad 
    \hat{p}(z_i) = \frac{1}{B} \sum_j^B \exp(\psi(z_i, z_j)).
\end{equation}
Substituting these estimators into \cref{eq:supcon_jensen} yields:
\begin{align}
    {\cal L}_\text{SupCon} &\geq - {\mathbb E}_i \left[ \log \left(\frac{\frac{1}{|{\cal S}_{y_i}|}\sum_{s \in {\cal S}_{y_i}}\exp(\psi(z_i, z_s))}{B \cdot \frac{1}{B}\sum_j^B \exp(\psi(z_i, z_j))} \right)\right] \\
    &= -{\mathbb E}_i \left[ \log \left( \frac{\hat{p}(z_i|y_i)}{B \cdot \hat{p}(z_i)} \right)\right] \\
    &= -{\mathbb E}_i \left[ \log  \frac{\hat{p}(z_i|y_i)}{\hat{p}(z_i)} - \log B\right] \\
    &=- {\mathbb E}_i \left[ \log \frac{\hat{p}(z_i|y_i)}{\hat{p}(z_i)}\right] + \log B \\
    &\approx- {\mathbb E}_i \left[ \log \frac{p(z_i|y_i)}{p(z_i)}\right] + \log B \\
    &= - I(Z;Y) + \log B.
\end{align}
Therefore, SupCon yields the following variational lower bound on the task-conditioned mutual information:
\begin{equation}
    I(Z;Y) \geq \log B - {\cal L}_\text{SupCon}.
\end{equation}
\hfill $\square$

\subsection{KL Divergence to vMF for Compactness-Based Information-Minimality}
\label{app:proof_kl_vmf}
To quantify the extent to which the learned representation $Z$ retains task-irrelevant information from the input $X$ given the downstream target $Y$, we consider the conditional mutual information $I(Z; X | Y)$. Under the Markov assumption $Y \to X \to Z$ (\cref{eq:multimodal_markov_chain}), this can be reformulated as follows:
\begin{align}
    I(Z;X|Y) &= {\mathbb E}_{p(z,x,y)} \left[\log \frac{p(z,x|y)}{p(z|y) p(x|y)} \right] \\
    &={\mathbb E}_{p(z,x,y)} \left[\log\frac{p(z|x,y) p(x|y)}{p(z|y) p(x|y)} \right] \\
    &={\mathbb E}_{p(z,x,y)} \left[\log \left( \frac{p(z|x)}{p(z|y)}\right)\right]  \qquad (\because Y \to X \to Z)\\  
    &={\mathbb E}_{p(x, y)} \left[ {\mathbb E}_{p(z|x)} \left[ \log \left( \frac{p(z|x)}{p(z|y)}\right)\right]\right] \\ 
    &= {\mathbb E}_{p(x,y)} \left[ D_\text{KL}\left( p(z|x) \parallel p(z|y) \right) \right].
    \label{eq:mi_kld_relation}
\end{align}
Thus, minimizing $I(Z; X | Y)$ reduces to minimizing the KL divergence between the conditional distributions $p(z | x)$ and $p(z | y)$.
However, directly computing this KL divergence is intractable due to the unknown forms of the underlying distributions. To address this, we approximate both $p(z | x)$ and $p(z | y)$ using von Mises–Fisher (vMF) distributions, motivated by the fact that the representations $f(x)$ are $\ell_2$-normalized and lie on the unit hypersphere after pretraining via InfoNCE in the specialization stage.
The vMF distribution over a unit vector $z \in \mathbb{R}^d$ is defined as:
\begin{equation}
    {\rm vMF}(z;\mu,\kappa) = C_d(\kappa) \exp(\kappa \mu^\top z).
\end{equation}
where $\mu$ is a unit vector denoting the mean direction, $\kappa$ is the concentration parameter, and $C_d(\kappa)$ is the normalization constant.
We approximate the conditional distributions using vMF distributions. Specifically, the instance-level distribution $p(z | x)$ is modeled as a vMF distribution centered at the sample-specific mean direction $\mu_x = f(x)$, where $f$ denotes the encoder mapping input $x$ to its $\ell_2$-normalized representation. In contrast, the class-level distribution $p(z | y)$ is approximated by a vMF distribution whose mean direction $\hat{\mu}_y$ is obtained by normalizing the empirical class mean vector, defined as $\mu_y= \frac{1}{|\mathcal{S}_y|} \sum_{s \in \mathcal{S}_y} z_s$, where $\mathcal{S}_y$ denotes the set of samples sharing the same class label $y$, as defined in \cref{app:proof_supcon}:
\begin{equation}
p(z|x) = {\rm vMF}(z;\mu_x,\kappa_x),\qquad 
p(z|y) = {\rm vMF}(z;\hat{\mu}_y,\kappa_y).
\end{equation}
The KL divergence between the two vMF distributions becomes:
\begin{align}
    D_{\rm KL}\left( p(z|x) \parallel p(z|y) \right) &= {\mathbb E}_{p(z|x)}\left[ \log \frac{p(z|x)}{p(z|y)}\right] \\
    &= {\mathbb E}_{p(z|x)}\left[ \log \frac{C_d(\kappa_x)\exp(\kappa_x \mu^\top_x z)}{C_d(\kappa_y)\exp(\kappa_y \hat{\mu}^\top_y z)}\right] \\
    &= {\mathbb E}_{p(z|x)}\left[ \log \left(\frac{C_d(\kappa_x)}{C_d(\kappa_y)}\right) + \kappa_x \mu^\top_x z - \kappa_y \hat{\mu}^\top_y z\right] \\
    &= \log \left(\frac{C_d(\kappa_x)}{C_d(\kappa_y)}\right) + \kappa_x{\mathbb E}_{p(z|x)}\left[ \mu^\top_x z\right] - \kappa_y{\mathbb E}_{p(z|x)}\left[ \hat{\mu}^\top_y z\right].
\end{align}
For a vMF distribution $\mathrm{vMF}(z;\mu,\kappa)$, the expected value is
$\mathbb{E}[z] = A_d(\kappa)\mu$, where $A_d(\kappa)$ denotes the mean resultant length of the vMF distribution. Therefore, we compute:
\begin{align}
    {\mathbb E}_{p(z|x)}\left[ \mu_x^\top z\right] &= \mu_x^\top {\mathbb E}_{p(z|x)}\left[z\right] = \mu_x^\top \left( A_d(\kappa_x)\mu_x\right) = A_d(\kappa_x) \left(\mu_x^\top \mu_x\right) = A_d(\kappa_x),\\
    {\mathbb E}_{p(z|x)}\left[ \hat{\mu}_y^\top z\right] &= \hat{\mu}_y^\top {\mathbb E}_{p(z|x)}\left[ z\right] = \hat{\mu}_y^\top \left( A_d(\kappa_x)\mu_x\right)= A_d(\kappa_x)\left( \mu_x^\top \hat{\mu}_y\right).
\end{align}
Plugging these in, we obtain the closed-form KL divergence:
\begin{equation}
    D_\text{KL}\left( p(z|x) \parallel p(z|y) \right) = \log \left(\frac{C_d(\kappa_x)}{C_d(\kappa_y)}\right) + \kappa_x A_d(\kappa_x) - \kappa_yA_d(\kappa_x)\left( \mu_x^\top \hat{\mu}_y\right).
\end{equation}
However, the above expression still involves $C_d(\kappa)$ and $A_d(\kappa)$, both of which are defined in terms of modified Bessel functions and are computationally expensive to evaluate. To make this expression more tractable, we follow the common simplification of assuming identical concentration parameters: $\kappa_x = \kappa_y = \kappa$. This assumption corresponds to modeling equal intra-class and instance-level concentration. This yields:
\begin{align}
    D_\text{KL}\left( p(z|x) \parallel p(z|y) \right) &= \log \left(\frac{C_d(\kappa)}{C_d(\kappa)}\right) + \kappa A_d(\kappa) - \kappa A_d(\kappa)\left( \mu_x^\top \hat{\mu}_y\right) \\ 
    &= \kappa A_d(\kappa) \left(1 -  \mu_x^\top \hat{\mu}_y\right).
\end{align}
As a result, the conditional mutual information admits the following proportional surrogate:
\begin{equation}
    {\mathbb E}_{p(x,y)} \left[ D_\text{KL}\left( p(z|x) \parallel p(z|y) \right)\right] \propto - {\mathbb E}_{p(x,y)} \left[ \mu_x^\top \hat{\mu}_y \right].
\end{equation}
Hence, minimizing $I(Z; X | Y)$ can be achieved via a tractable surrogate that maximizes the cosine similarity between each sample’s direction vector and its class-conditional mean direction.
\hfill $\square$
\section{Auxiliary Losses}
\label{app:auxiliary_losses}
To ensure stable and meaningful expert utilization in the Mixture-of-Experts (MoE) architecture, we employ four auxiliary losses that guide the router $g$ to assign tokens effectively and encourage specialization across experts. Since the MoE router adopts top-$k$ sparse routing, it is prone to training instabilities such as expert collapse and assignment imbalance. These issues hinder effective use of model capacity and compromise expert-level diversity and specialization. 
To mitigate these challenges, we incorporate and adapt several strategies proposed in prior works~\cite{shazeer2017, riquelme2021scaling, mustafa2022multimodal} to fit our framework. Specifically, our auxiliary loss is defined as a weighted sum of four components:
\begin{equation}
    {\cal L}_\text{aux} = \lambda_\text{imp} {\cal L}_\text{imp} + \lambda_\text{load}{\cal L}_\text{load} + \lambda_\text{local}{\cal L}_\text{local} + \lambda_\text{global}{\cal L}_\text{global}.
\end{equation}
Each component is defined based on token-level routing statistics within a training batch. Let ${\cal B} = \{1, \ldots, B \}$ denote the set of batch indices, and let the intermediate representation of each sample be given by a sequence $[{\mathbf x}_{b,t}]_{t=1}^{T_b}$, where ${\mathbf x}_{b,t} \in {\mathbb R}^{D_\text{model}}$. Here, $T_b$ denotes the sequential length of the input sample $b$. We define the set of all tokens in the batch as ${\mathbf X} = \{{\mathbf x}_{b,t} \mid b \in {\cal B}, \ t = 1, \ldots, T_b \}$.
\subsection{Importance Loss}
The importance loss encourages balanced utilization of experts by enforcing that the total soft routing weight assigned to each expert is approximately uniform across the batch. From a soft-routing perspective, this objective mitigates the tendency of routing probabilities to concentrate excessively on a small subset of experts.
Formally, the importance of expert $i$ is defined as
\begin{equation}
    {\rm Imp}_i({\mathbf X}) = \sum_{{\mathbf x} \in {\mathbf X}} {\rm softmax}({\mathbf W}_g {\mathbf x})_i,
\end{equation}
where ${\mathbf W}_g$ denotes the linear projection associated with the router. If the resulting importance vector ${\rm Imp}({\mathbf X}) = \{ {\rm Imp}_i({\mathbf X}) \}_{i=1}^{N_\text{expert}}$ is highly imbalanced, routing weights may collapse onto a small number of experts, leading to degraded specialization and inefficient use of model capacity.
To prevent this behavior, we penalize the dispersion of the importance distribution by minimizing its coefficient of variation:
\begin{equation}
    {\cal L}_\text{imp} = \left( \frac{{\rm std(Imp({\mathbf X}))}}{{\rm mean(Imp({\mathbf X}))}} \right)^2 \propto {\rm var({\rm Imp}({\mathbf X}))}.
\end{equation}
\subsection{Load Loss}
While the importance loss promotes balanced soft routing weights, it does not guarantee balanced hard assignments due to the nature of top-$k$ selection. In practice, even with a uniform importance distribution, the router may repeatedly assign tokens to a small subset of experts, leaving others unused. Load loss directly addresses this assignment imbalance by encouraging equalized expert selection.
However, since the number of assigned tokens is a discrete and non-differentiable quantity, we adopt a probabilistic proxy following \citet{shazeer2017, riquelme2021scaling}. Specifically, we estimate the probability that expert $i$ is selected in the top-$k$ for a given token ${\mathbf x}_{b,t}$.
During the forward pass, we add Gaussian noise to the expert scores, yielding ${\mathbf W}_g {\mathbf x}_{b,t} + \epsilon$, where $\epsilon \sim {\cal N}(0, \sigma^2)$ and $\sigma = \frac{1}{N_\text{expert}}$ controls the noise scale. The top-$k$ threshold is then defined as the $k$-th largest value among the noisy scores:
\begin{equation}
    {\rm threshold}_k({\mathbf  x}_{b,t})=\max_{k\text{-}{\rm th}}({\mathbf W}_g{\mathbf x}_{b,t} + \epsilon).
\end{equation}
To compute the selection probability of expert $i$, we re-sample an independent noise term $\epsilon_\text{new} \sim {\cal N}(0, \sigma^2)$ and estimate the probability that expert $i$'s score exceeds the top-$k$ threshold:
\begin{align}
    p_i({\mathbf x}) &= P(({\mathbf W}_g {\mathbf x}_{b,t})_i + \epsilon_\text{new} \geq {\rm threshold}_k({\mathbf x}_{b,t})) \\
    &= P(\epsilon_\text{new} \geq {\rm threshold}_k({\mathbf x}_{b,t}) - ({\mathbf W}_g {\mathbf x}_{b,t})_i).
\end{align}
Using these probabilities, we define the expected load of expert $i$ over the batch as:
\begin{align}
    {\rm load}_i({\mathbf X}) &= \sum_{{\mathbf x} \in {\mathbf X}} p_i({\mathbf x}), \\
    {\rm load}({\mathbf X})&=\{ {\rm load}_i({\mathbf X})\}_{i=1}^{N_\text{expert}}.
\end{align}
Finally, to encourage balanced load across experts, we minimize the squared coefficient of variation of the load distribution:
\begin{equation}
    {\cal L}_\text{load} = \left( \frac{{\rm std(load({\mathbf X}))}}{{\rm mean(load({\mathbf X}))}} \right)^2 \propto {\rm var}({\rm load}({\mathbf X})).
\end{equation}
\subsection{Local Entropy Loss}
In each MoE layer, the router produces a probabilistic distribution over experts for every input token. For a given token ${\mathbf x}_{b,t}$, the expert routing distribution is defined as
\begin{equation}
    p({\rm expert} \mid {\mathbf x}_{b,t}) \in {\mathbb R}^{N_\text{expert}},
\end{equation}
which represents the relative contribution or routing strength of each expert in processing the token. The top-$k$ operation subsequently selects a subset of experts based on this distribution.
The local entropy loss is defined as the average entropy of the token-level routing distributions across the batch:
\begin{equation}
    {\cal L}_{\rm local} = {\mathbb E}_{{\mathbf x} \in {\mathbf X}} \left[ H(p({\rm expert} \mid {\mathbf x})) \right],
\end{equation}
where $H(\cdot)$ denotes the standard Shannon entropy. Minimizing ${\cal L}_{\rm local}$ encourages lower entropy in the routing distributions, thereby promoting more confident and concentrated assignments of each token to a small number of experts.
This sharpening of token-level routing encourages consistent and decisive expert selection, which in turn strengthens expert-level specialization by reducing overlap among expert responsibilities.

\subsection{Global Entropy Loss}
While the local entropy loss encourages sharp and confident expert selection at the token level, the global entropy loss promotes diversity in expert utilization from a batch-level perspective. To quantify the overall distribution of expert usage, we compute the average routing distribution across all tokens in the batch:
\begin{equation}
    \tilde{p}({\rm experts}) = {\mathbb E}_{{\mathbf x} \in {\mathbf X}} \left[ p({\rm expert} \mid {\mathbf x}_{b,t}) \right] \in {\mathbb R}^{N_\text{expert}},
    \label{eq:expert_marginal_distribution}
\end{equation}
where $\tilde{p}({\rm experts})$ represents the marginal expert utilization across the batch. It reflects the average routing wight received by each expert. If this distribution is highly skewed toward a few experts, it may lead to expert collapse and underutilization of model capacity.
To mitigate this issue, the global entropy loss encourages the marginal distribution to approach uniformity by maximizing its entropy. Specifically, it is defined as:
\begin{equation}
    {\cal L}_{\rm global} = - H \bigl(\tilde{p}_m({\rm experts})\bigr).
\end{equation}

\section{Experimental Details}
\label{app:experimental_details}

\subsection{Dataset Description}
\label{app:dataset_description}

We conduct experiments on four representative benchmarks provided by the real-world multimodal benchmark suite, MultiBench~\cite{liang2021multibench}. MultiBench covers a wide range of multimodal tasks across domains and provides pre-extracted modality-specific features along with standardized evaluation protocols. Following prior works~\cite{liang2023factorized,wang2025an}, we adopt the same input features and evaluation settings for fair comparison. The benchmarks used in our experiments are as follows:
\vspace{-0.15cm}
\begin{enumerate}
    \item \textsc{MOSEI}~\cite{bagher-zadeh-etal-2018-multimodal}: A sentiment and emotion recognition benchmark comprising approximately 23,000 monologue video segments. Each sample is labeled with a sentiment intensity score in the range of $[-3, 3]$. Following \citet{liang2023factorized} and \citet{wang2025an}, we convert the scores into binary labels (positive/negative) and use the provided vision and text features. \vspace{-0.05cm}
    \item \textsc{MOSI}~\cite{zadeh2016mosi}: A sentiment analysis benchmark similar to MOSEI, consisting of 2,199 short video clips from YouTube. Sentiment scores are similarly binarized, and vision and text features are used. \vspace{-0.05cm}
    \item \textsc{UR-FUNNY}~\cite{hasan-etal-2019-ur}: A humor detection benchmark built from TED Talk segments, containing over 16,000 samples. Each sample is labeled for the presence or absence of humor, using vision and text modalities. \vspace{-0.05cm}
    \item \textsc{MUStARD}~\cite{castro-etal-2019-towards}: A sarcasm detection benchmark consisting of 690 video clips from TV shows such as Friends, The Golden Girls, and The Big Bang Theory. The task is framed as binary classification using pre-extracted vision and text features. \vspace{-0.1cm}
\end{enumerate}
\vspace{-0.15cm}
We exclude the \textsc{MIMIC-III}~\cite{PhysioNet-mimiciii-1.4} benchmark from our experiments. This dataset comprises time-series signals and tabular metadata from over 40,000 ICU patients. Prior works adopt GRUs for the time-series modality and MLPs for the tabular modality. In contrast, our method employs Transformer-based Mixture-of-Experts encoders across all modalities. Due to this architectural mismatch and the data structure of MIMIC-III, we exclude it from evaluation.

\subsection{Implementation Details}

We design our experiments to follow the settings of prior works~\cite{liang2023factorized,wang2025an} for fair comparison. For all benchmarks, we use the same data splits provided by MultiBench. Each modality is processed by an independent Mixture-of-Experts encoder, and the modality-specific feature encoders consist of 5 layers, following prior work. The expansion ratio $\rho$ is fixed to 8 across all datasets, while the granularity $\chi$ and preservation ratio $p$ are varied across experiments.
All models are trained end-to-end from random initialization. After representation learning, performance is evaluated using linear probing by training a separate linear classifier. Following~\citet{wang2025an}, we report the mean and standard deviation of downstream prediction accuracy over three random seeds.

\section{Additional Experimental Results}
\label{app:additiona_experimental_results}

\subsection{Entropy-Based Monitoring of Router Behavior at Selection Stage}
\label{app:router_entropy_at_selection}

\begin{figure}[!t]
    \centering
    \includegraphics[width=0.92\textwidth]{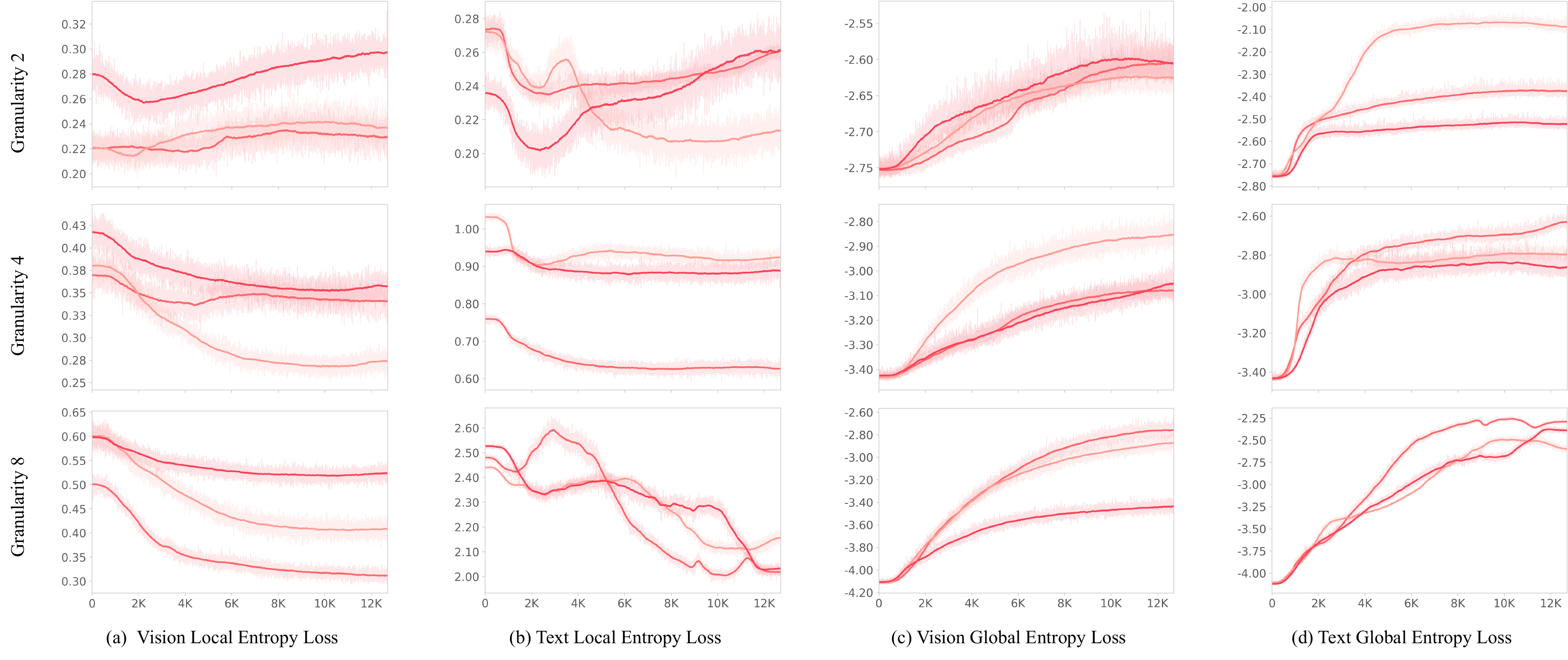}
    \vspace{-0.2cm}
    \caption{Entropy-based monitoring of router behavior on \textsc{MOSEI} during the Selection stage. We visualize the dynamics of local and global entropy losses for both vision and text modalities across different granularity levels ($\chi=2,4,8$).}
    \vspace{-0.05cm}
    \label{fig:mosei_entropy}
\end{figure}

\begin{figure}[!t]
    \centering
    \includegraphics[width=0.92\textwidth]{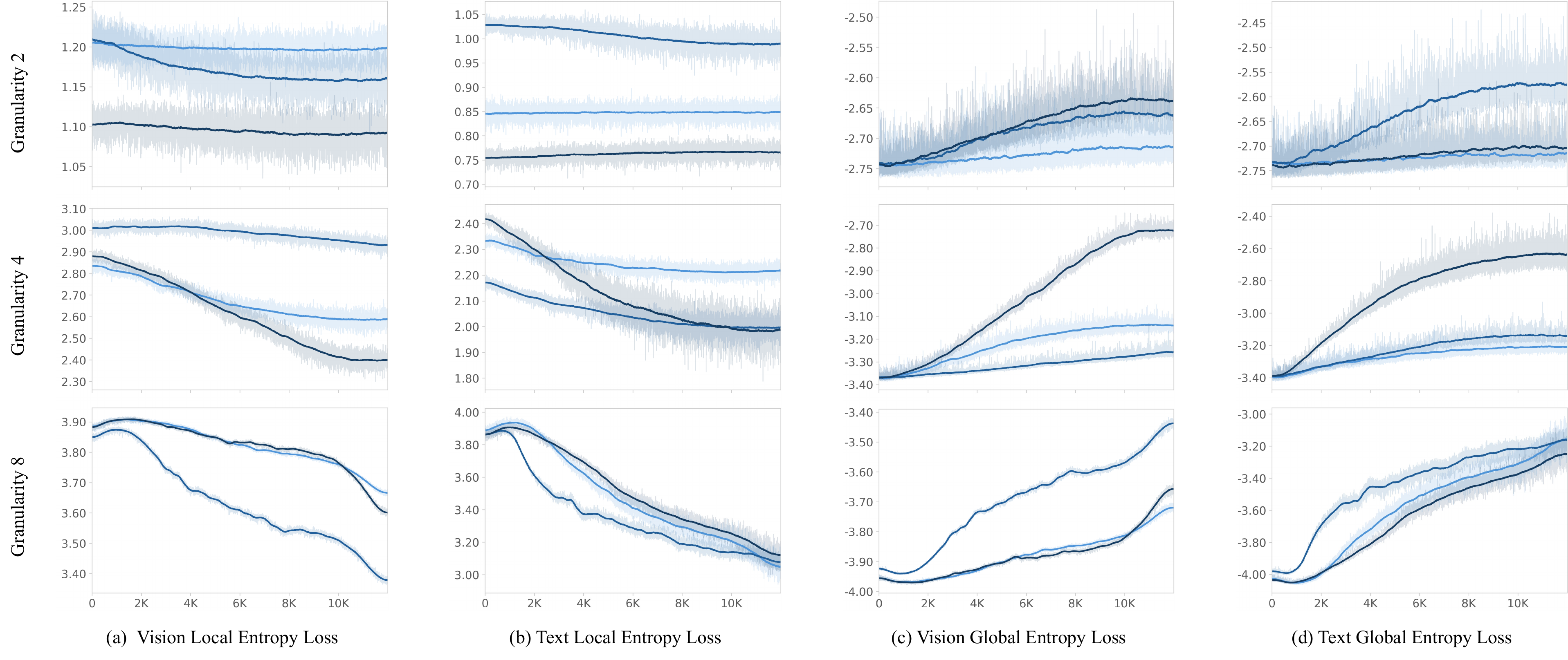}
    \vspace{-0.2cm}
    \caption{Entropy-based monitoring of router behavior on \textsc{MOSI} during the Selection stage. We visualize the dynamics of local and global entropy losses for both vision and text modalities across different granularity levels ($\chi=2,4,8$).}
    \vspace{-0.05cm}
    \label{fig:mosi_entropy}
\end{figure}

\begin{figure}[!t]
    \centering
    \includegraphics[width=0.92\textwidth]{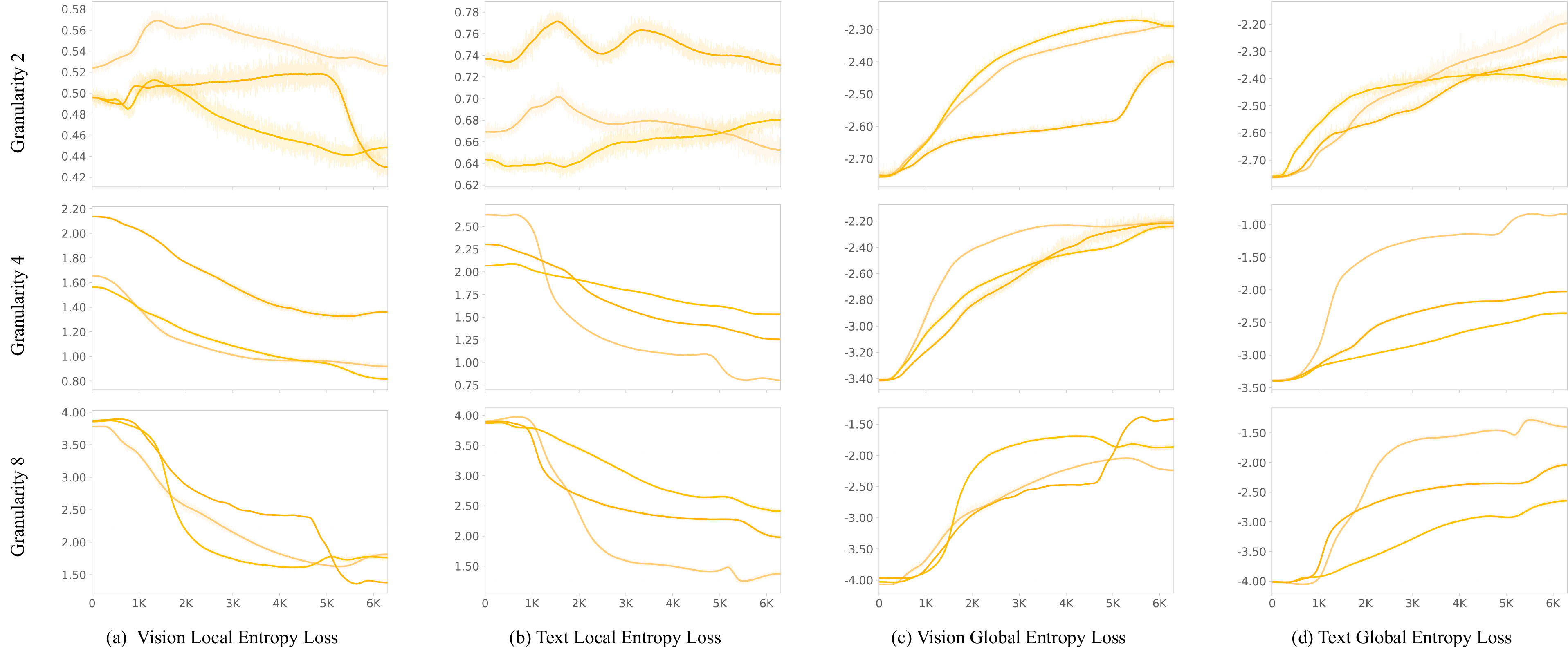}
    \vspace{-0.2cm}
    \caption{Entropy-based monitoring of router behavior on \textsc{UR-FUNNY} during the Selection stage. We visualize the dynamics of local and global entropy losses for both vision and text modalities across different granularity levels ($\chi=2,4,8$).}
    \vspace{-0.05cm}
    \label{fig:ur-funny_entropy}
\end{figure}

\begin{figure}[!t]
    \centering
    \includegraphics[width=0.92\textwidth]{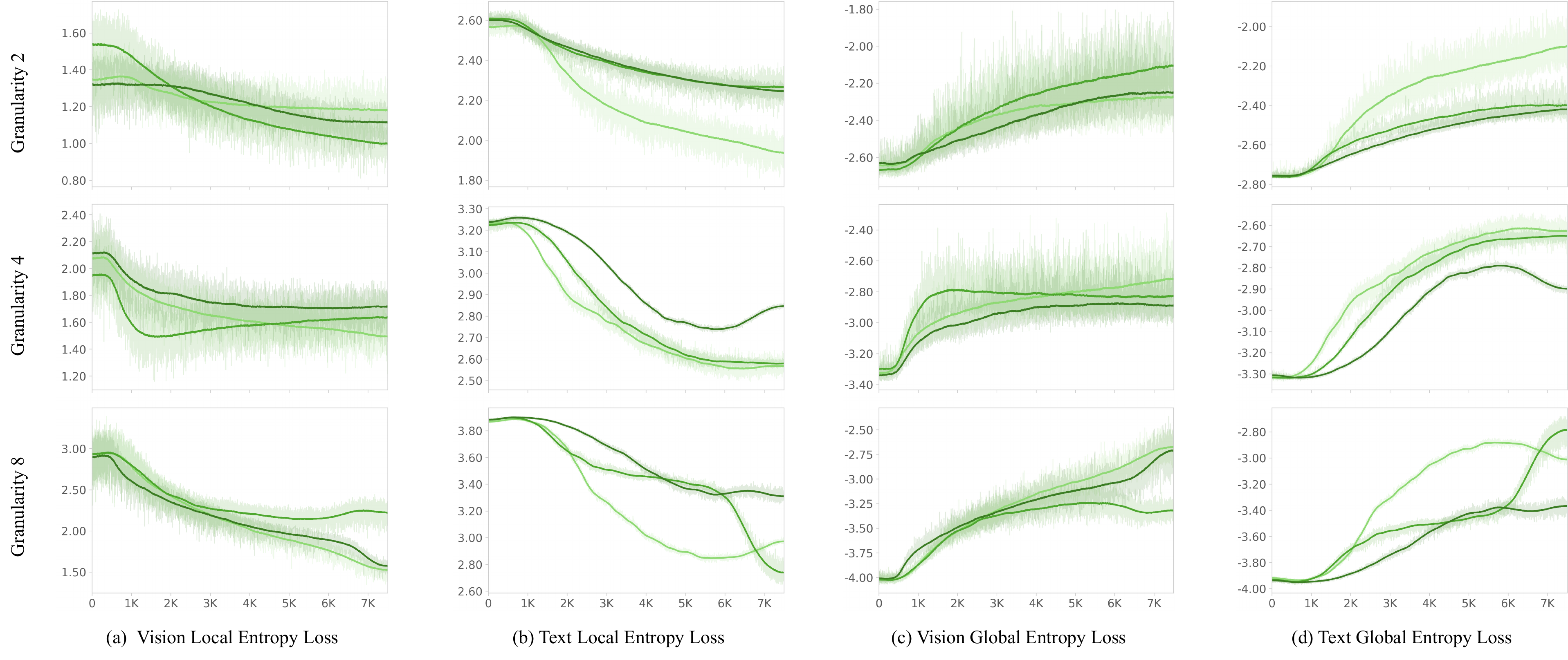}
    \vspace{-0.2cm}
    \caption{Entropy-based monitoring of router behavior on \textsc{MUStARD} during the Selection stage. We visualize the dynamics of local and global entropy losses for both vision and text modalities across different granularity levels ($\chi=2,4,8$).}
    \vspace{-0.05cm}
    \label{fig:mustard_entropy}
\end{figure}

During the Selection stage, the router is optimized to satisfy Task-Sufficiency and Information-Minimality by maximizing mutual information among label-consistent samples. While the attention and expert modules are frozen, only the lightweight router is fine-tuned.
Although auxiliary losses are not used as training objectives in this stage, we monitor them throughout training to qualitatively analyze the behavior of the router. Specifically, we track both the local entropy loss (measuring per-token expert selection confidence) and the global entropy loss (measuring expert usage diversity across the batch), as defined in \cref{app:auxiliary_losses}.
\cref{fig:mosei_entropy,fig:mosi_entropy,fig:ur-funny_entropy,fig:mustard_entropy} show the dynamics of local and global entropy for both modalities (vision and text) across different granularity settings ($\chi=2,4,8$) for each dataset. 

We observe that higher granularity leads to a decrease in local entropy, indicating more confident and deterministic expert selection over time. In contrast, at low granularity ($\chi=2$), local entropy tends to decrease more slowly, fluctuate, or even increase, suggesting routing ambiguity and less reliable expert discrimination. These trends support our analysis in \cref{subsec:analisys_of_S3_stages_on_MOSEI}: at low granularity, each expert captures a broad mixture of semantic concepts due to the small number of large experts, resulting in greater semantic entanglement. This makes it more difficult for the router to identify task-relevant experts.

Furthermore, we observe that entropy stability is strongly influenced by dataset scale. Larger datasets such as MOSEI and UR-FUNNY exhibit lower per-step fluctuation. In contrast, smaller datasets such as MOSI and MUStARD show greater per-step variance, indicating higher instability in expert selection during training.
This observation aligns with the variance patterns discussed in \cref{subsec:comparison_with_prior_methods}, where smaller datasets were shown to induce higher performance variability due to limited data. The entropy-based monitoring thus provides further evidence that dataset scale impacts not only downstream accuracy but also the stability of the router's expert selection process.

Finally, we note a consistent decrease in global entropy (i.e., increase in negative entropy) during training, reflecting the router’s tendency to converge toward a more selective usage of task-relevant experts. This outcome aligns with the label-consistency assumption underlying our Selection stage: samples with the same label are expected to activate semantically aligned experts, leading to sharper and more focused routing distributions.

\begin{wraptable}{r}{0.52\linewidth}
\vspace{-0.8\baselineskip}
\centering
\begin{center}
\resizebox{\linewidth}{!}{
\renewcommand{\arraystretch}{1.2}
\begin{tabular}{C{0.8cm}C{0.8cm}|C{2.0cm}C{2.0cm}C{2.0cm}C{2.0cm}}
\Xhline{3.5\arrayrulewidth}
${\cal L}_\text{suff}$ & ${\cal L}_\text{min}$ & \textbf{\textsc{MOSEI}} & \textbf{\textsc{MOSI}} & \textbf{\textsc{UR-FUNNY}} & \textbf{\textsc{MUStARD}} \\
\hline
 &  & 75.78(0.32) & 63.56(2.18) & 63.61(0.49) & 58.70(3.16) \\
\hline
\cmark & \cmark & \underline{77.36(0.29)} & 64.29(1.27) & 63.52(0.62) & \textbf{61.59(0.72)} \\
\cmark &  & \textbf{77.53(0.50)} & \underline{65.31(0.76)} & \underline{64.08(0.57)} & \underline{61.11(5.44)} \\
 & \cmark & 77.07(0.31) & \textbf{66.67(1.31)} & \textbf{64.18(0.74)} & 60.63(3.42) \\
\Xhline{3.5\arrayrulewidth}
\end{tabular}
}
\end{center}
\caption{Ablation of router fine-tuning losses ($\chi=8$). Prediction accuracy (\%) on four benchmarks, averaged over three seeds (std. in parentheses). The first row corresponds to Specialization only, the others apply Selection with different loss combinations.}
\label{tab:ablation_study}
\vspace{-0.4cm}
\end{wraptable}

\subsection{Ablation Study}

\paragraph{Selection Stage: Loss Combination Ablation.}
We analyze the individual and combined contributions of the two objectives used in the Selection stage: the Task-Sufficiency loss ($\mathcal{L}_\text{suff}$) and the Information-Minimality loss ($\mathcal{L}_\text{min}$). 
\Cref{tab:ablation_study} reports results under the granularity setting $\chi=8$, where only the router is fine-tuned on top of a model pretrained with the Specialization stage, using different combinations of the two losses.

The first row corresponds to applying Specialization only, without the Selection stage, and yields consistently lower performance across all benchmarks. This indicates that, despite the structural expressiveness of the pretrained latent space, task-specific performance remains limited without explicit task-adaptive selection.
The remaining rows show results obtained by fine-tuning the router using $\mathcal{L}_\text{suff}$ only, $\mathcal{L}_\text{min}$ only, and their combination, respectively. In all datasets, applying the Selection stage leads to consistent performance improvements over the Specialization-only baseline.

Notably, although the two losses operate through different mechanisms, they result in comparable performance gains. 
The Task-Sufficiency loss ($\mathcal{L}_\text{suff}$) emphasizes task-relevant experts by promoting information sharing among label-consistent samples, while the Information-Minimality loss ($\mathcal{L}_\text{min}$) suppresses task-irrelevant experts by discouraging label-invariant variations.
Despite these differing formulations, both objectives ultimately serve a common goal: highlighting task-relevant semantic components.
The fact that each loss independently yields meaningful improvements supports the effectiveness of our design in inducing task-adaptive expert selection from complementary perspectives.

\paragraph{Applying Sparsification without the Selection Stage.}

\begin{figure}[!t]
    \centering
    \includegraphics[width=0.92\textwidth]{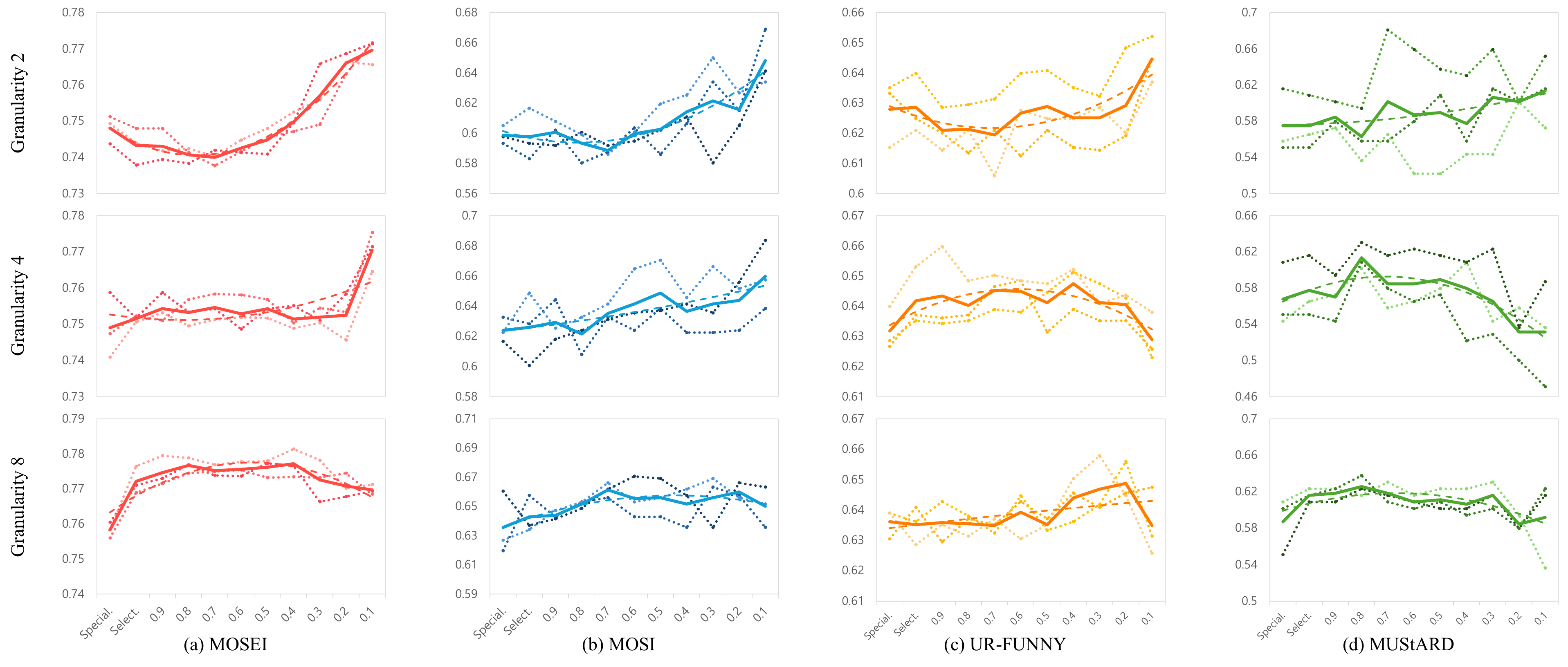}
    \vspace{-0.2cm}
    \caption{Performance across four benchmarks for $\chi \in \{2,4,8\}$. Solid lines show the mean over three random seeds, dashed lines their trend, and dotted lines individual seeds. All results follow the full S3 pipeline, with $p$ progressively decreased during Sparsification.}
    \vspace{-0.05cm}
    \label{fig:main_results}
\end{figure}

\begin{figure}[!t]
    \centering
    \includegraphics[width=0.92\textwidth]{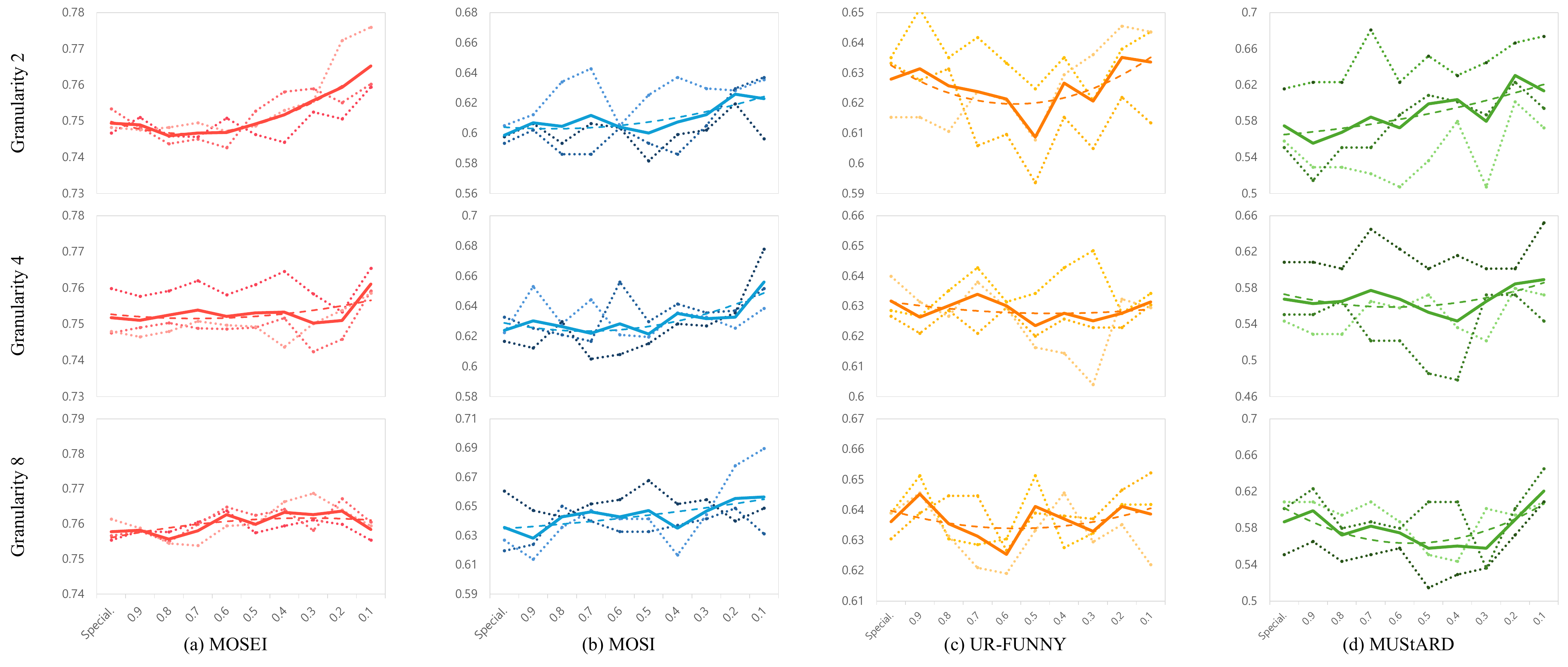}
    \vspace{-0.2cm}
    \caption{Performance when Sparsification is applied directly after Specialization, without the Selection stage. Results are shown for $\chi \in \{2,4,8\}$ using the same y-axis scale as \cref{fig:main_results}.}
    \vspace{-0.05cm}
    \label{fig:ablation_selection_stage}
\end{figure}

We further examine the effect of applying Sparsification without the Selection stage, i.e., performing pruning directly after Specialization.
\Cref{fig:ablation_selection_stage} visualizes the resulting pruning curves, using the same y-axis ranges and resolution as \Cref{fig:main_results} for direct comparison with the full S3 framework.

Overall, when Sparsification is applied without prior Selection, performance changes do not exhibit consistent or monotonic trends.
This behavior arises because the router operates independently of task in this setting.
Without explicitly identifying and emphasizing task-relevant experts, pruning cannot reliably distinguish between semantically important and unimportant routing paths, leading to structurally unpredictable performance variations.

These observations indicate that Sparsification is not an independent performance-improvement mechanism, but rather relies on the task-adaptive routing structure established during the Selection stage.
Meaningful performance-efficiency trade-offs emerge only after task-relevant expert activations have been aligned.
When applied in isolation, Sparsification fails to fully realize its structural benefits.

In summary, this ablation study demonstrates that the three stages of the S3 framework are not independently interchangeable.
Instead, stable and interpretable task-adaptive sparsity emerges from their sequential structure: Specialization, Selection, Sparsification.
Each stage provides the structural foundation for the next, and Sparsification becomes effective only after expert routing has been aligned with task semantics through the Selection stage.

\clearpage

\begin{wraptable}{r}{0.5\linewidth}
\vspace{-0.8\baselineskip}
\centering
\begin{center}
\resizebox{\linewidth}{!}{
\renewcommand{\arraystretch}{1.2}
\begin{tabular}{L{3.0cm}C{2.0cm}C{2.0cm}C{2.0cm}}
\Xhline{3.5\arrayrulewidth}
 & & $\chi$ & \\
\cmidrule(r){2-4}
\textbf{\textsc{Dataset}} & 2 & 4 & 8 \\
\hline
\textbf{\textsc{MOSEI}} & 0.0984 & 0.2075 & 0.4574 \\
\textbf{\textsc{MOSI}} & 0.0984 & 0.2075 & 0.4574 \\
\textbf{\textsc{UR-FUNNY}} & 0.0982 & 0.2071 & 0.4565 \\
\textbf{\textsc{MUStARD}} & 0.2059 & 0.4538 & 1.0708 \\
\Xhline{3.5\arrayrulewidth}
\end{tabular}
}
\end{center}
\caption{Ratio of trainable parameters during router fine-tuning in the Selection stage (\%). We report the proportion of trainable parameters relative to the total model parameters for different granularity settings ($\chi=2,4,8$). Only the router is updated, with attention and expert weights frozen.
}
\label{tab:trained_parameter_during_finetuning}
\vspace{-0.4cm}
\end{wraptable}

\subsection{Trainable Parameter Ratio in Selection Stage}
\label{app:computational_cost}

To quantify the lightweight nature of the Selection stage, we report the ratio of trainable parameters involved in router fine-tuning. During this stage, all attention and expert layers are frozen, and only the router parameters are updated. \Cref{tab:trained_parameter_during_finetuning} presents the proportion of trainable parameters relative to the total model size for each granularity setting ($\chi=2,4,8$).
As shown in the table, the router accounts for less than 1\% of the total parameters in all cases, with values ranging from 0.0984\% to 1.0708\%. These results confirm that our Selection stage enables highly efficient fine-tuning with negligible parameter updates.

\subsection{Sparsification Efficiency Analysis}
\label{app:sparsification_parameter_utilization}

To analyze the computational efficiency of our sparsification stage, we measure the proportion of active parameters per token under varying preservation ratios. For each preservation ratio $p\in[0,1]$, we compute the expected fraction of activated parameters per token, normalized to 100\% at $p=1$.
As shown in \cref{tab:effect_of_each_components}, reducing the preservation ratio leads to a consistent decrease in active parameters across all datasets and granularities. These results validate the practicality of our sparsification stage as an inference-time efficiency control mechanism without requiring any additional training.

\begin{table*}[t]
\setlength{\tabcolsep}{1.4pt}
\centering
\resizebox{\textwidth}{!}{
\renewcommand{\arraystretch}{1.3}
\begin{tabular}{L{2.7cm}|C{0.9cm}|C{2.15cm}C{2.15cm}C{2.15cm}C{2.15cm}C{2.15cm}C{2.15cm}C{2.15cm}C{2.15cm}C{2.15cm}}
\Xhline{3.5\arrayrulewidth}
\textbf{\textsc{Dataset}} & $\chi$ & 0.9 & 0.8 & 0.7 & 0.6 & 0.5 & 0.4 & 0.3 & 0.2 & 0.1 \\
\hline
\multirow{3}{*}{\textbf{\textsc{MOSEI}}} & 2 & 96.61 & 93.23 & 89.84 & 86.45 & 83.07 & 79.68 & 76.30 & 72.91 & 69.52 \\
& 4 & 96.61 & 93.21 & 89.82 & 86.43 & 83.04 & 79.64 & 76.25 & 72.86 & 69.46 \\
& 8 & 96.59 & 93.19 & 89.78 & 86.38 & 82.97 & 79.57 & 76.16 & 72.75 & 69.35 \\
\hline
\multirow{3}{*}{\textbf{\textsc{MOSI}}} & 2 & 96.61 & 93.22 & 89.83 & 86.44 & 83.05 & 79.66 & 76.27 & 72.88 & 69.49 \\
& 4 & 96.60 & 93.21 & 89.81 & 86.41 & 83.02 & 79.62 & 76.22 & 72.82 & 69.43 \\
& 8 & 96.59 & 93.18 & 89.77 & 86.36 & 82.95 & 79.54 & 76.13 & 72.72 & 69.31 \\
\hline
\multirow{3}{*}{\textbf{\textsc{UR-FUNNY}}} & 2 & 96.70 & 93.41 & 90.11 & 86.82 & 83.52 & 80.23 & 76.93 & 73.64 & 70.34 \\
& 4 & 96.70 & 93.40 & 90.10 & 86.79 & 83.49 & 80.19 & 76.89 & 73.59 & 70.29 \\
& 8 & 96.69 & 93.37 & 90.06 & 86.74 & 83.43 & 80.11 & 76.80 & 73.48 & 70.17 \\
\hline
\multirow{3}{*}{\textbf{\textsc{MUStARD}}} & 2 & 96.87 & 93.74 & 90.60 & 87.47 & 84.34 & 81.21 & 78.08 & 74.94 & 71.81 \\
& 4 & 96.86 & 93.71 & 90.57 & 87.42 & 84.28 & 81.13 & 77.99 & 74.85 & 71.70 \\
& 8 & 96.83 & 93.66 & 90.49 & 87.32 & 84.15 & 80.99 & 77.82 & 74.65 & 71.48 \\
\Xhline{3.5\arrayrulewidth}
\end{tabular}
}
\vspace{0.1cm}
\caption{Expected per-token parameter activate (\%) across preservation ratios $p\in [0,1]$ during the Sparsification stage. Values are normalized to 100\% at $p=1$. The metric reflects the average fraction of active parameters. The metric reflects the average fraction of active parameters involved in semantic transformation, attention, and expert computation, excluding the router.
}
\label{tab:effect_of_each_components}
\end{table*}

\clearpage

\end{document}